\documentclass{article}
\usepackage[utf8]{inputenc}

\title{Sequential Neural Posterior and Likelihood Approximation}
\author{Samuel Wiqvist$^{1}$, Jes Frellsen$^{2, \star}$, Umberto Picchini$^{3, \star}$}
\date{\footnotesize{$^{1}$Centre for Mathematical Sciences, Lund University, Lund, Sweden \\
$^{2}$Department of Applied Mathematics and Computer Science,
Technical University of Denmark, Denmark \\
$^{3}$Department of Mathematical Sciences, Chalmers University of Technology and the University of Gothenburg, Gothenburg, Sweden}}

\PassOptionsToPackage{numbers, compress}{natbib}

\usepackage{natbib}
\usepackage{graphicx}

\usepackage{amsmath}
\usepackage{amsfonts}       
\usepackage{amssymb}
\usepackage{xcolor}
\usepackage[margin=2.5cm]{geometry}
\usepackage{mathtools}
\usepackage{commath}
\usepackage{booktabs} 

\usepackage{microtype}
\usepackage{graphicx}
\usepackage{subfigure}
\usepackage{booktabs} 
\usepackage{hyperref}
\usepackage{graphicx}

\usepackage{booktabs} 
\usepackage{arydshln}
\usepackage{commath}
\usepackage{aligned-overset}

\usepackage[textwidth=20mm,textsize=scriptsize]{todonotes}
\usepackage[ruled,vlined,linesnumbered]{algorithm2e}

\setcitestyle{authoryear}

\newcommand\blfootnote[1]{%
  \begingroup
  \renewcommand\thefootnote{}\footnote{#1}%
  \addtocounter{footnote}{-1}%
  \endgroup
}

\begin{document}

\addtocontents{toc}{\protect\setcounter{tocdepth}{0}}


\maketitle

\begin{abstract}
We introduce the \textit{sequential neural posterior and likelihood approximation} (SNPLA) algorithm. SNPLA is a normalizing flows-based algorithm for inference in implicit models, and therefore is a simulation-based inference method that only requires simulations from a generative model. SNPLA avoids Markov chain Monte Carlo sampling and correction-steps of the parameter proposal function that are introduced in similar methods, but that can be numerically unstable or restrictive. By utilizing the reverse KL divergence, SNPLA manages to learn both the likelihood and the posterior in a sequential manner. Over four experiments, we show that SNPLA performs competitively when utilizing the same number of model simulations as used in other methods, even though the inference problem for SNPLA is more complex due to the joint learning of posterior and likelihood function. Due to utilizing normalizing flows  SNPLA generates posterior draws much faster (4 orders of magnitude) than MCMC-based methods.  
\blfootnote{$^{\star}$Equal contribution.}
\blfootnote{Contact: \href{mailto:samuel.wiqvist@matstat.lu.se}{samuel.wiqvist@matstat.lu.se}; \href{mailto:jefr@dtu.dk}{jefr@dtu.dk}; \href{mailto:picchini@chalmers.se}{picchini@chalmers.se}.}
\blfootnote{Code: \href{https://github.com/SamuelWiqvist/snpla}{https://github.com/SamuelWiqvist/snpla}.}
\end{abstract}

\section{Introduction}
\label{intro}

Simulation-based inference (SBI) refers to methods that allow for inference in implicit statistical models, meaning that the likelihood function is only known implicitly via simulations from a generative model. In this work we introduce the \textit{sequential neural posterior and likelihood approximation} (SNPLA), a SBI algorithm for Bayesian inference that bypasses expensive Markov chain Monte Carlo (MCMC) sampling by efficiently generating draws from an approximate posterior distribution. Additionally, and unlike other similar SBI methods, SNPLA learns a computationally cheap approximation of the likelihood function, thus allowing simulations from this ``learned'' likelihood to be efficiently performed.

Traditionally, approximate Bayesian computation (ABC) \citep{beaumont2002approximate,marin2012approximate} is the most popular methodology for inference in implicit models. Other important methods include: synthetic likelihoods (SL) \citep{wood2010statistical,price2018bayesian}, Bayesian optimization \citep{gutmann2016bayesian}, classification based methods such as likelihood-free inference by ratio estimation (LFIRE) \citep{thomas2020likelihood}, and pseudomarginal methods \citep{andrieu2009pseudo,andrieu2010particle}. More recent methods are reviewed by \citet{cranmer2019frontier} and have been benchmarked in \citet{lueckmann2101benchmarking}. Also, in recent years normalizing flows \citep{kobyzev2020normalizing} have turned especially popular in simulation-based inference \citep{papamakarios19snl, greenberg19aautopost, radev2020bayesflow}, due to the ease of probabilistic sampling and density evaluation \citep{papamakarios2019normalizing}.  
Some of the methods that are particularly relevant for our work are: \textit{sequential neural posterior estimation} (SNPE) (SNPE-A \citep{papamakarios2016fast}, SNPE-B \citep{lueckmann2017flexible}, and SNPE-C \citep{greenberg19aautopost}); \textit{sequential neural likelihood estimation} (SNL) \citep{papamakarios19snl}; \textit{sequential neural ratio estimation} (SNRE-A \citep{hermans2019likelihood}, and SNRE-B \citep{durkan2020contrastive}); and BayesFlow \citep{radev2020bayesflow}. 

Our work will focus in particular on SNL and SNPE, since SNPLA is inspired by both. The main disadvantage of SNPE is the \textit{correction step} that must be used so that SNPE learns the correct posterior distribution. SNL avoids this correction step (since SNL learns the likelihood model). However, SNL relies on MCMC for sampling from the posterior, which is time-consuming and restricts which posteriors can reasonably be learned with success, since MCMC exploration of complex surfaces (e.g.\@ multi-modal targets) can be challenging. Our \textbf{main contribution} is the proposed SNPLA method which addresses both these issues. Namely, (i) SNPLA avoids the correction step in SNPE by utilizing  the reverse Kullback–Leibler (KL) divergence, and (ii) SNPLA bypasses the typically expensive MCMC runs by using normalizing flows to model both an approximate posterior and an approximate likelihood, resulting in efficient sampling from both. Empirically, we show that SNPLA is on average almost $10^4$ times faster than SNL in producing posterior samples (Table \ref{tab:res_hyp_param}).
\textbf{Additionally}, it can be highly valuable to access the normalizing flow-based likelihood model, learned as a by-product of (ii), since this model approximates the data generating process. Thus by sampling from the learned likelihood model, one can rapidly generate artificial data from an approximate generative model. Of course, the latter will be most informative for input parameters that are similar to those that generated the observed data set $x_{\text{obs}}$. An example of the usefulness of learning the likelihood model is illustrated in Section \ref{sec:HH}. \textbf{Finally}, we also show that it is possible to simultaneously learn summary statistics of the data, altogether with the likelihood model and the posterior model, thus providing a flexible plug-and-play inference framework. The \textbf{code for replicating the results} can be found at \href{https://github.com/SamuelWiqvist/snpla}{https://github.com/SamuelWiqvist/snpla}.

\section{Simulation-based inference for implicit models}
\label{simbasedinference}

The implicit statistical model is given by 
\begin{equation*}
    \theta \sim p(\theta), \,\, x \sim p(x | \theta),
\end{equation*}
where $p(x | \theta)$ is the likelihood function associated to generic data $x$ (hence is sometimes denoted ``global likelihood''), and whose functional form we assume unknown. However, we assume the likelihood to be implicitly encoded via an associated computer \textit{simulator} that allows us to generate artificial data, conditionally on an input given by some arbitrary parameter $\theta$ and a stream of pseudorandom numbers (and possibly additional covariates or inputs that we do not explicitly represent in our notation). The parameter prior $p(\theta)$ specifies our a-priori beliefs regarding $\theta$. Implicit models are flexible since they only require us to specify a simulator and not the functional form of the likelihood. 

As motivated in the Introduction, we are going to focus on SNPE and SNL. SNPE directly learns an approximation $\tilde{p}_{\phi_P}(\theta | x_{\text{obs}})$ to the parameter posterior conditionally on observed data $x_{\text{obs}}$, while SNL learns an approximation $\tilde{p}_{\phi_L}(x | \theta)$ of the global likelihood. The models are parameterized with weights $\phi_P$ and $\phi_L$ respectively. However, the global likelihood model $\tilde{p}_{\phi_L}(x | \theta)$ is trained with data influenced by the observed data set $x_{\text{obs}}$. This means that $\tilde{p}_{\phi_L}(x | \theta)$ will be most accurate for values of $\theta$ having high density under the posterior $p(\theta | x_{\text{obs}})$.  

Both SNPE and SNL are sequential schemes that are made data-efficient by employing a proposal distribution $\hat{p}(\theta | x_{\text{obs}})$ that is sequentially adapted to leverage more information from the most recent approximation of the posterior. The SNPE and the SNL schemes are outlined in the supplementary material. For SNPE we have that the proposal distribution is corrected with the factor $p(\theta) / \hat{p}(\theta | x_{\text{obs}})$. This \textit{correction step} is necessary to ensure that the newly constructed parameter proposal is valid, see \citet{papamakarios2016fast} for details. The correction step can either be done analytically \citep{papamakarios2016fast}, numerically \citep{lueckmann2017flexible}, or via reparameterization \citep{greenberg19aautopost}.
The correction step can introduce complexities into SNPE. For example, the closed-form correction of \citet{papamakarios2016fast} can be numerically unstable (if the proposal prior has higher precision than the estimated conditional density) and is restricted to Gaussian and uniform proposals, limiting both the robustness and flexibility of the approach. Regarding the correction in \citet{lueckmann2017flexible}, the introduction of importance weights greatly increases the variance of parameter updates during learning, which can lead to slow or inaccurate inference \cite{greenberg19aautopost}. For SNL this correction is not necessary since SNL learns the likelihood model $\tilde{p}_{\phi_L}(x | \theta)$. On the other hand, SNL uses MCMC to sample $\theta \sim \hat{p}(\theta | x_{\text{obs}})$, which is time-consuming. For instance, our analysis shows (see right sub-table of Table \ref{tab:res_hyp_param}) that SNPLA generates posterior samples on average $12{,}000$ times more rapidly than SNL. Also, MCMC sampling can be unfeasible for some targets with complex geometries.

\section{Sequential neural posterior and likelihood approximation}
\label{sec:method}

Here we detail our proposed method. The main idea of the SNPLA method is to jointly learn both an approximation $\tilde{p}_{\phi_P}(\theta | x_{\text{obs}})$ of the parameter posterior, and an approximation $\tilde{p}_{\phi_L}(x | \theta)$ of the likelihood function. Thus SNPLA has two learnable models: 
\begin{enumerate}
    \item \textbf{Parameter posterior model} $\tilde{p}_{\phi_P}(\theta | x_{\text{obs}})$, approximating the parameter posterior distribution $p(\theta | x_{\text{obs}})$.   

    \item \textbf{Likelihood model $\tilde{p}_{\phi_L}(x | \theta)$}. Since we are considering an implicit statistical model, we consider the likelihood model $\tilde{p}_{\phi_L}(x | \theta)$ as approximating the data generating process $p(x | \theta)$.  
\end{enumerate}  

Both the posterior model and the likelihood model are parameterized via normalizing flows, with weights $\phi_P$ and $\phi_L$ respectively. The use of normalizing flows is critical since these can be trained using either the forward or the reverse KL divergence \citep{papamakarios2017masked}. The relevant properties and notation for normalizing flows used for SNPLA are presented in the supplementary material.

For SNPLA the likelihood model $\tilde{p}_{\phi_L}(x | \theta)$ is learned via training data sampled from a proposal distribution $\hat{p}(\theta | x_{\text{obs}})$. However, the obtained likelihood approximation $\tilde{p}_{\phi_L}(x | \theta)$ is also used to train the posterior model $\tilde{p}_{\phi_P}(\theta | x_{\text{obs}})$, so that we jointly learn both the posterior and the likelihood. The SNPLA scheme is outlined in Algorithm \ref{algo:snpla}.

\begin{algorithm}[ht]
\caption{SNPLA}
\label{algo:snpla}
\DontPrintSemicolon 
\KwIn{Untrained likelihood model $\tilde{p}_{\phi_L}(x | \theta)$, untrained posterior model $\tilde{p}_{\phi_P}( \theta | x)$, number of iterations $R$, number of training samples per iteration $N$, number of training samples per iteration for the posterior model $N_P$, decay rate $\lambda>0$. 
}
\KwOut{Trained likelihood model $\tilde{p}_{\phi_L}(x | \theta)$, trained posterior model $\tilde{p}_{\phi_P}( \theta | x)$.}

Set  $\mathcal{D_L}=\{\emptyset\}$.

\For{$r = 1:R$} {
    \tcc{Step 1: Update likelihood model with training data sampled from a mixture of the prior and the current posterior model}
    Sample for $n=1:N$ $$(\theta_n,x_n) \sim p(x| \theta)\hat{p}_r(\theta | x_{\text{obs}}),$$
    where $\hat{p}_r(\theta | x_{\text{obs}}) = \alpha p(\theta) + (1-\alpha) \tilde{p}_{\phi_P}(\theta | x_{\text{obs}})$ and, for example, $\alpha = \exp(-\lambda \cdot (r-1))$.

    Update training data $\mathcal{D_L} = [\theta_{1:N}, x_{1:N}]\cup\mathcal{D_L}$. \\

    Update $\tilde{p}_{\phi_L}(x | \theta)$ by minimizing the following loss
    $$
    \mathcal{L}(\phi_L) = -E_{\hat{p}(\theta,x| x_{\text{obs}})}\big[\log\tilde{p}_{\phi_L}(x | \theta) \big] \propto E_{\hat{p}(\theta  | x_{\text{obs})}}\Big[D_{KL}\big(p(x | \theta) ) \big|\big| \tilde{p}_{\phi_L}(x | \theta) \big)\Big]. 
    $$
    \If{$r = 1$}{
        \tcc{Step 2': Hot-start for learning the posterior model}
       Using the prior-predictive samples $[\theta_{1:N},  x_{1:N}]$, update the posterior model by minimizing the following loss 
        $$\mathcal{L}(\phi_P) \propto -E_{p(\theta,x) = p(x|\theta)p(\theta)}\big[\log \tilde{p}_{\phi_P}(\theta | x) \big].$$
    }
     \tcc{Step 2: Update the posterior model with training data generated from the current posterior model}
     \For{$j = 1:N_P/N_{\mathrm{mini}}$}{
        For $i = 1:N_{\mathrm{mini}}$: Sample $\theta_i \sim \tilde{p}_{\phi_P}(\theta | x_{\text{obs}})$
        
        Update posterior model, i.e. obtain a new $\phi_P$ by \@ minimizing the loss (reverse KL divergence):
        $$
        \mathcal{L}(\phi_P) = D_{KL}\big(\tilde{p}_{\phi_P}(\theta | x_{\text{obs}})  \big|\big| \tilde{p}_{\phi_L}(x_{\text{obs}}|\theta)p(\theta)\big). 
        $$
    }
}
\end{algorithm}

\paragraph{Motivation for the construction}
Assume that $\tilde{p}_{\phi_P}(\theta | x_{\text{obs}})$ and $\tilde{p}_{\phi_L}(x | \theta)$ are universal approximators, and that we run SNPLA for one iteration (i.e.\@ $R=1$) without the hot-start procedure. In this optimal case, after training in \textbf{step 1}, we will have that 
\begin{align*}
E_{\hat{p}(\theta  | x_{\text{obs})}}\Big[D_{KL}\big(p(x | \theta) ) \big|\big| \tilde{p}_{\phi_L}(x | \theta) \big)\Big] = 0 \Longrightarrow \tilde{p}_{\phi_L}(x | \theta) = p(x | \theta).
\end{align*}
Thus we have now fully learnt the global likelihood, since training data is generated from the prior-predictive distribution at iteration one. Subsequently, after training in \textbf{step 2}, we will have that 
\begin{align*}
D_{KL}\big(\tilde{p}_{\phi_P}(\theta | x_{\text{obs}})  \big|\big| \tilde{p}_{\phi_L}(x_{\text{obs}}|\theta)p(\theta)\big) = 0 \Longrightarrow \tilde{p}_{\phi_P}(\theta | x_{\text{obs}}) &\propto \tilde{p}_{\phi_L}(x_{\text{obs}}|\theta)p(\theta) \\
\overset{\text{optimal case}}&{=} 
p(x_{\text{obs}}|\theta)p(\theta)
\end{align*}
Thus, in the optimal case, we learn the true global likelihood and the true parameter posterior. Of course, in practice, we will utilize models with limited capacity, and we will also need to run SNPLA for several iterations to leverage informative training data for the observed data set $x_{\text{obs}}$ that we are considering. 

\paragraph{Properties}

In step 1, the likelihood model $\tilde{p}_{\phi_L}(x | \theta)$ is updated with data generated via the proposal distribution $\hat{p}(\theta | x_{\text{obs}})$, where the latter is set to sequentially leverage more information from the posterior model $\tilde{p}_{\phi_P}(\theta | x_{\text{obs}})$. The parameter $\lambda$ governs how rapidly we want to leverage information from the posterior model.  In practice, we have found it useful to use a rather high $\lambda \approx 0.7-0.9$, so that the proposal distribution is quickly adapted. For more information regarding the choice of $\lambda$ see Section \ref{sec:sens_analysis}. $N$ governs the number of training data points used in this step. However, since we need to run $N$ simulations in step 1 it is advantageous to keep $N$ conservative, particularly if the model simulator is slow. 

In step 2' the posterior model is updated with samples from the prior-predictive distribution. This step is included on pragmatic grounds since we found SNPLA to exhibit convergence problems if this step is not included. Step 2' acts as a pre-training step of the posterior model $\tilde{p}_{\phi_P}(\theta | x_{\text{obs}})$ such that the posterior model is set to learn its true target in the first iteration of the algorithm. 

Finally, in step 2, the posterior model is updated with samples generated from the current version of the posterior model. These samples are generated via a \textit{simulation-on-fly} scheme where each mini-batch is simulated from the most recent version of the posterior model. This means that in step 2 we loop over the number of mini-bathes $N_P/N_{\mathrm{mini}}$ ($N_{\mathrm{mini}}$ being the number of samples in one mini-batch, and $N_P$ the total number of samples), and for each mini-batch we simulate new training data. Thus, each mini-batch used in step 2 is unique, since it is simulated from the most recent posterior approximation. Since the training data in step 2 is generated from the flow model, we can take $N_P$ to be much larger than $N$, and typically one order of magnitude larger. 

The posterior model is challenging to learn since it is set to target an approximation of its true target. Furthermore, the learning of the posterior model can be sensitive to potentially catastrophic moves in the weight space, since each mini-batch is generated on-the-fly. However, our experience also shows that the learning of the likelihood model is considerably easier. 
The learning process of the posterior model has been made more robust by utilizing several strategies: one component of these strategies is to include the hot-start approach in step 2' (something that we already discussed above). We have also found it to be useful to use a large batch size in step 2 ($N_{\mathrm{mini}} \approx 1000$) since a large batch size smooths the training process and thus avoids potentially catastrophic moves.
We have also seen that the convergence problems can be addressed by carefully selecting and tuning the ADAM optimizer's learning rate.  We have obtained the best results when using a moderate to large learning rate for the posterior model during the first few iterations, which is then sequentially decreased. This allows the posterior model to rapidly explore the weights space in the early iterations and find a reasonable approximation to the posterior. When decreasing the learning rate, we avoid large catastrophic moves of the posterior model's weights, while allowing the posterior model to be fine-tuned when accessing more data.  

\subsection{Learning the summary statistics} \label{sec:snpla_learn_summary_stats}
So far, we denoted with $\tilde{p}(\theta | x_{\text{obs}})$ the posterior model that SNPLA learns.
However, we can also condition on some function $S(\cdot)$ of the data. In that case, we obtain the following model 
\begin{align} \label{eq:post_model_s}
    \tilde{p}_{\phi_P}(\theta | S_{\phi_{P_S}}(x_{\text{obs}})),
\end{align}
where $\phi_{P_{S}}$ denotes weights that are specific to the $S(\cdot)$ function, since in our work we assume $S(\cdot)$ parameterized with some neural network.
We can interpret the function $S(\cdot)$ as mapping the data $x_{\text{obs}}$ into a set of summary statistics of the data $S(x_{\text{obs}})$. It is possible to automatically learn $S(\cdot)$ altogether with the likelihood model and the posterior model, as we show in section \ref{sec:mvg}, thus providing a general and flexible learning framework. Considering \eqref{eq:post_model_s} can be particularly useful if $x_{\text{obs}}$ is high-dimensional and/or contains some spatial or temporal structure. Thus we want the network $S_{\phi_{P_{S}}}(\cdot)$ to leverage the features of the data $x_{\text{obs}}$ and therefore, for exchangeable data, we could for instance use a DeepSets network \cite{zaheer2017deep} (as we do in section \ref{sec:mvg}), and for Markovian time-series data it is possible to use a partially exchangeable network \cite{wiqvist2019pen}. The network $S_{\phi_{P_{S}}}(\cdot)$ is trained jointly with the posterior model in \eqref{eq:post_model_s}. So, when incorporating trainable summary statistics, line 6 and 9 of Algorithm \ref{algo:snpla} is modified so that the posterior model and the summary statistics network are updated according to the following loss 
\begin{align*}
\mathcal{L}(\phi_P, \phi_{P_{S}}) =  D_{KL}\big(\tilde{p}_{\phi_{P}}(\theta |S_{\phi_{P_{S}}}(x_{\text{obs}}))  \big|\big|  \tilde{p}_{\phi_L}(x_{\text{obs}}|\theta)p(\theta)\big).
\end{align*}
Importantly, notice that the summary statistics network $S_{\phi_{P_{S}}}(\cdot)$ is not included in the likelihood model $\tilde{p}_{\phi_L}(x | \theta)$. The likelihood model still learns a model of the full data set $x_{\mathrm{obs}}$, and not a model for the set of summary statistics computed by $S_{\phi_{P_{S}}}(\cdot)$. However, for complex data with spatial or temporal structures $\tilde{p}_{\phi_L}(x_{\text{obs}}|\theta)$ can be set up so that the these data features are leveraged in the likelihood model. For instance, specialized flow models for images, audio, and text data have been developed \cite{papamakarios2019normalizing}, and these can be used in the likelihood model to leverage the data features.

\section{Experiments} \label{experiments}
We consider the following case-studies: a multivariate  Gaussian (MV-G) model, the two-moons (TM) model \citep{greenberg19aautopost}, the Lotka-Volterra (LV) model, and the Hodgkin-Huxley (HH) model. The full experimental setting is presented in the supplementary material. 

\subsection{Proof-of-concept: Multivariate Gaussian}\label{sec:mvg}
We consider the following conjugate MV Gaussian example from \citet{radev2020bayesflow}
\begin{align*}
\mu &\sim N(\mu  | \mu_{\mu}, \Sigma_{\mu}), \qquad x \sim N(x | \mu, \Sigma),
\end{align*}
where both Gaussians have dimension two.
The covariance $\Sigma$ is assumed to be known, so the main goal here is to infer the posterior for the mean $p(\mu | x_{\text{obs}})$. The posterior $p(\mu | x_{\text{obs}})$ is analytically known and  we evaluate the quality of the inference by computing the KL divergence between the analytical and the approximate posterior, same as in \cite{radev2020bayesflow}, for details see the supplementary material. 

We consider three versions of the MV Gaussian study: (i) ``five observations'', where data $x_{\text{obs}}$ consists of five two-dimensional samples, and the likelihood model $\tilde{p}_{\phi_L}(x | \theta)$ is set to directly target these five observations; (ii) ``summary statistics'' where data are given by five summary statistics 
obtained from a two-dimensional vector of $100$ observations, and the likelihood model is set to target the summary statistics of these data; and finally (iii)  ``learnable summary statistics'', where we use a small DeepSets network \cite{zaheer2017deep} to automatically learn the summary statistics following the method in Section \ref{sec:snpla_learn_summary_stats} based on five observations. We ran all methods independently for 10 times (each time with a different set of observed data), using $R = 10$ iterations with $N = 2{,}500$ model simulations for each iteration. For SNPLE we used $N_P = 40{,}000$ (``five observations'' and ``learnable summary statistics''), and $N_P = 10{,}000$. (``summary statistics''). For SMC-ABC, however, we utilized up to, in total, $N = 10^6$ model simulations. 

\begin{figure}[tb]
\begin{center}
\subfigure[MV-G ``summary statistics'']{\includegraphics[width=.45\columnwidth]{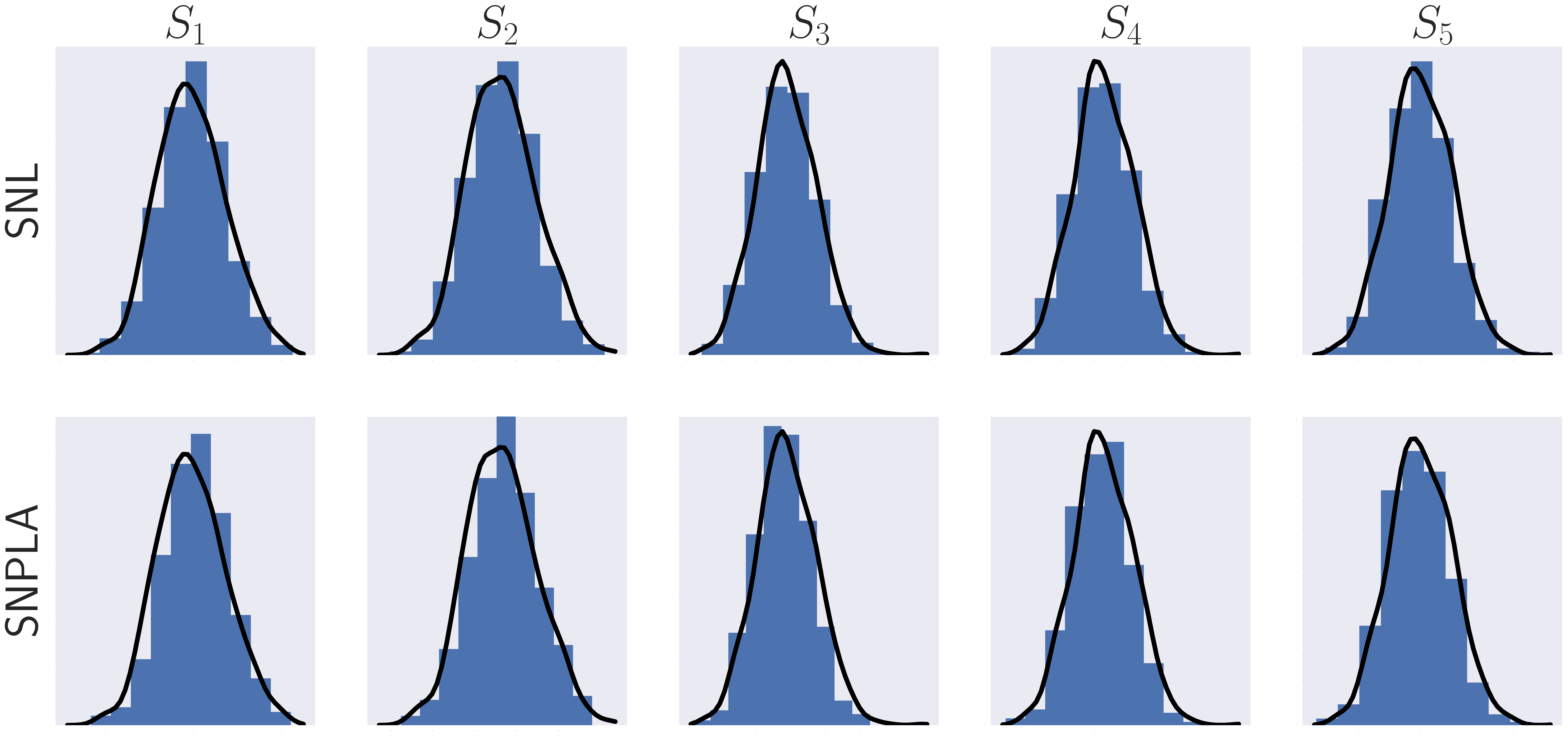} \label{fig:mv_gauss_post_pred}}
\subfigure[TM]{\includegraphics[width=.45\columnwidth]{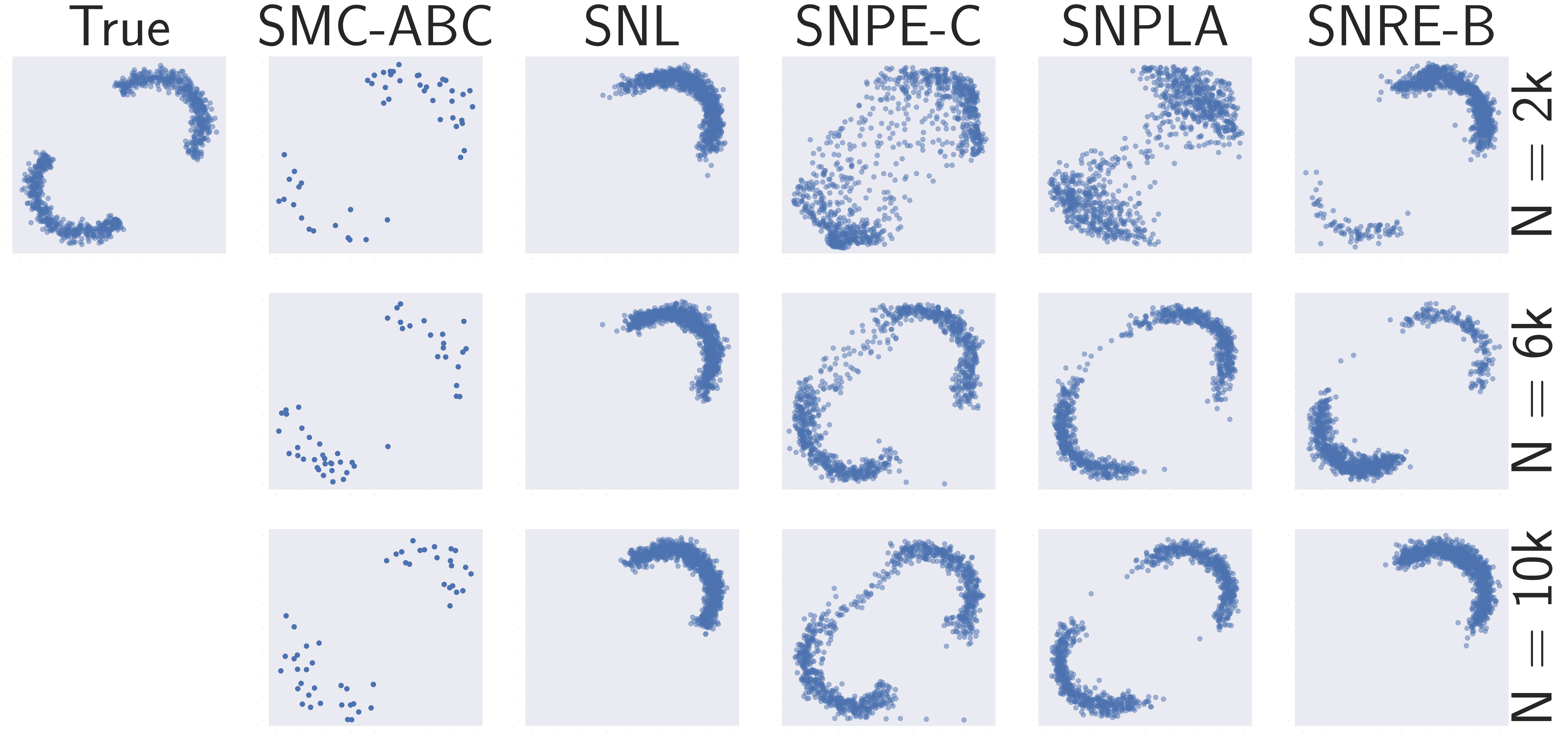} \label{fig:TM_post_samples}}
\caption{MV-G and TM: (a) Marginal densities of summaries from the true likelihood model (solid lines) and simulated summary statistics from the learned likelihood model (histograms), produced conditionally on parameters $\theta$ from the true posterior; 
(b) Samples from the approximate posterior distributions and the exact one. Results from one of the ten runs. Results from the first iteration (top), from the third iteration (middle) and the fifth iteration (bottom). } 
\end{center}
\end{figure}

\begin{figure}[tb]
\begin{center}
\subfigure[``Five observations'']{\includegraphics[width=.3\columnwidth]{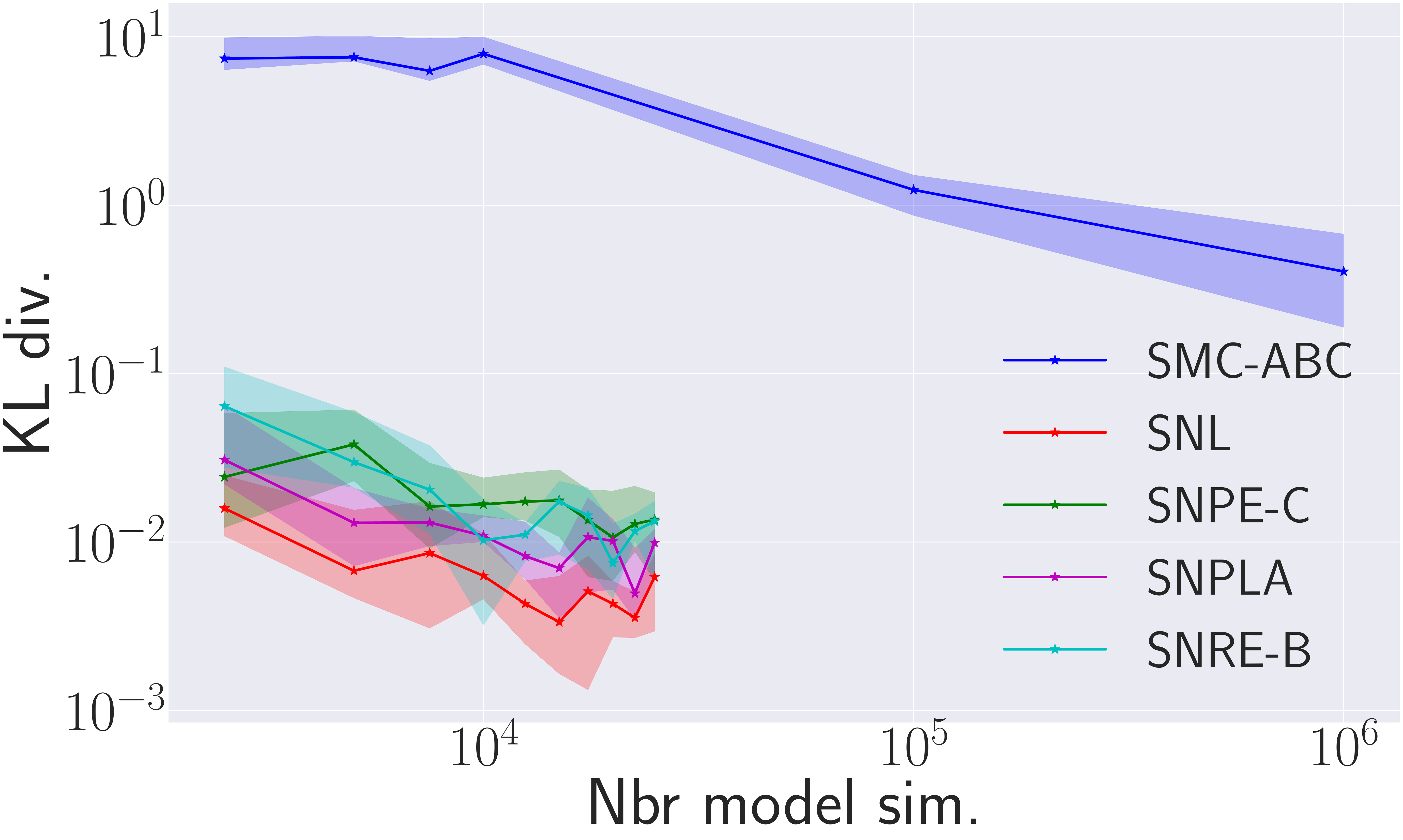} \label{fig:mv_gauss_post_inferenc_res_five}}
\subfigure[``Summary statistics'']{\includegraphics[width=.3\columnwidth]{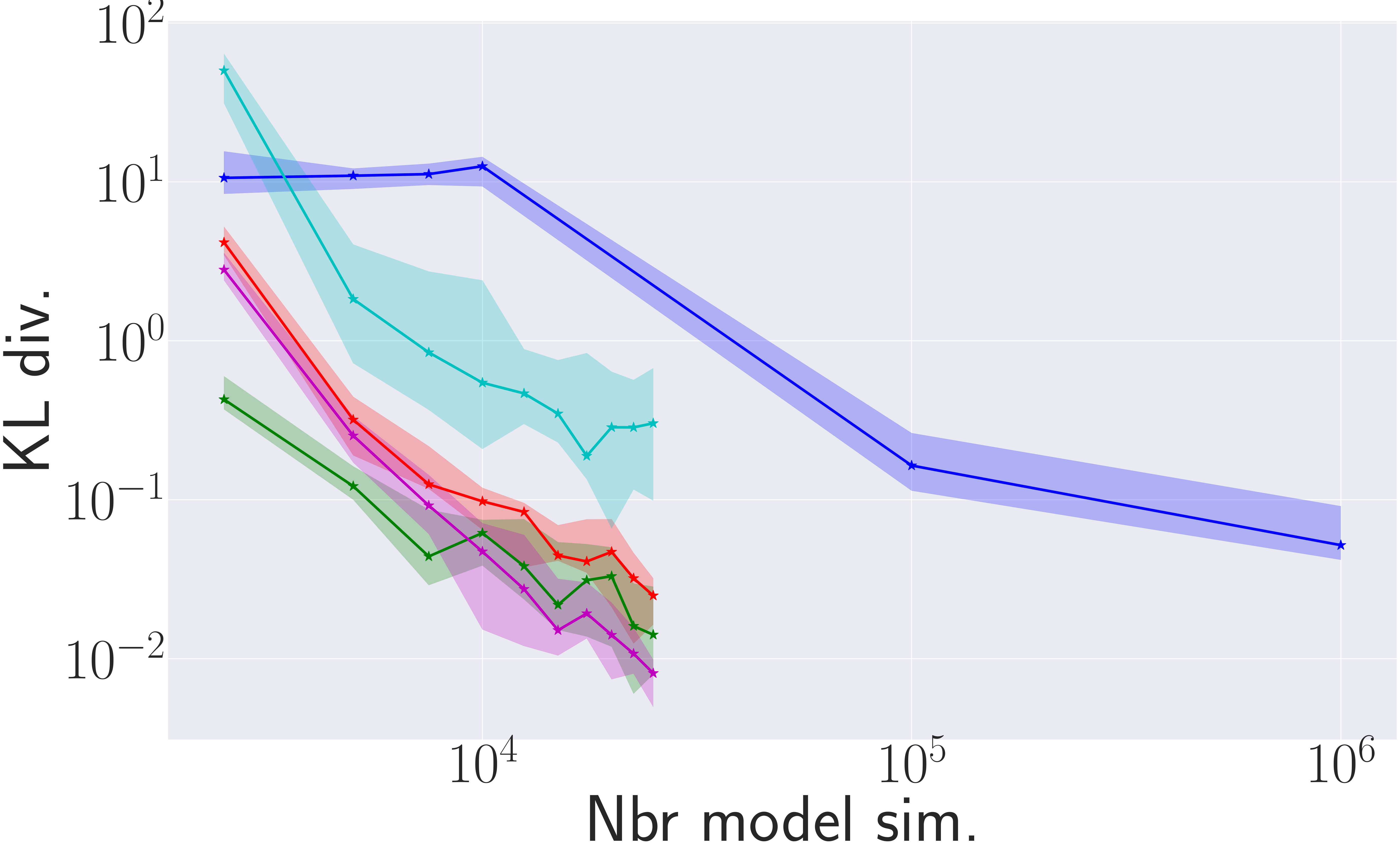} \label{fig:mv_gauss_post_inferenc_res_ss}}
\subfigure[``Learnable summary statistics'']{\includegraphics[width=.3\columnwidth]{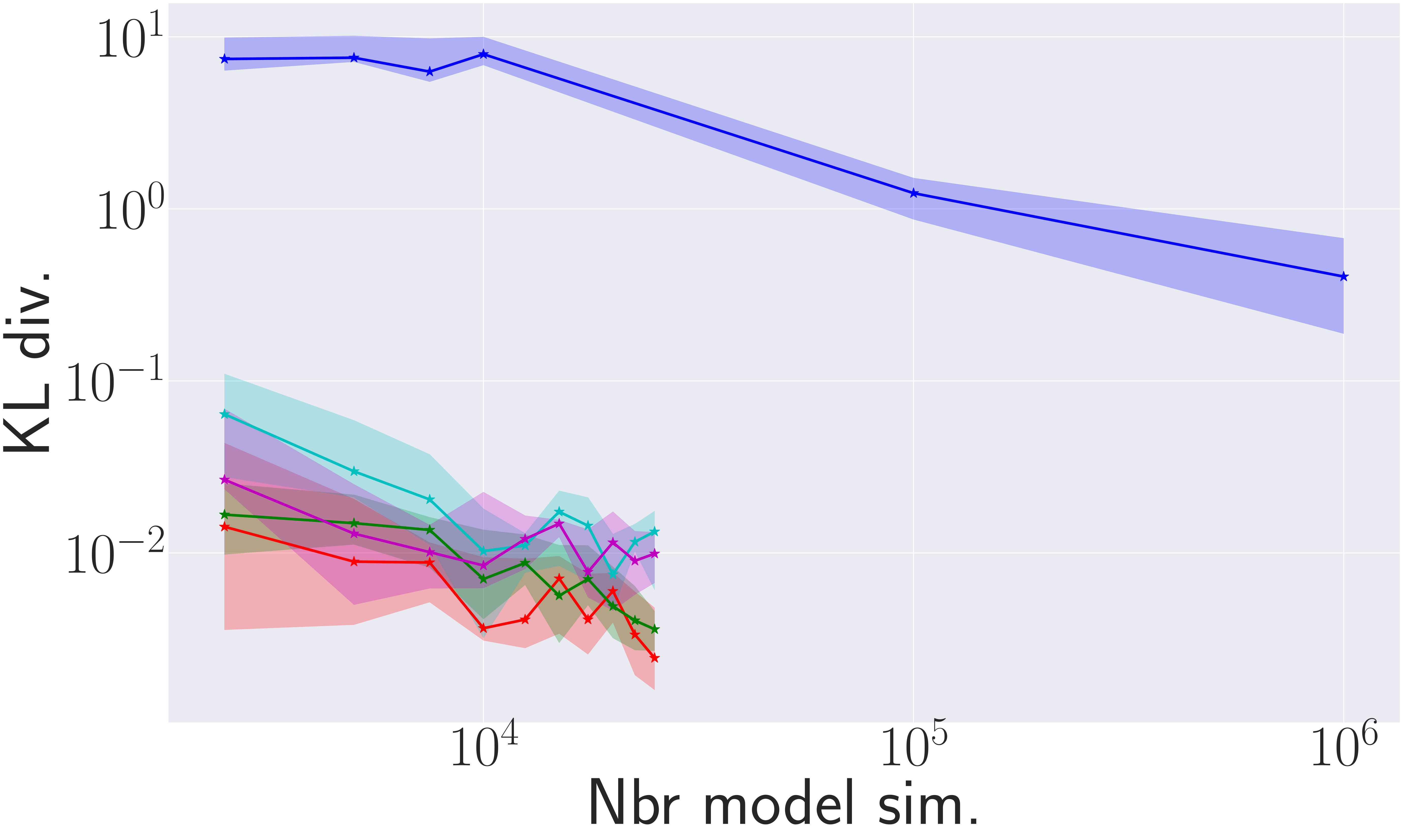} \label{fig:mv_gauss_post_inferenc_res_ls}}
\\
\subfigure[Two-moons]{\includegraphics[width=.3\columnwidth]{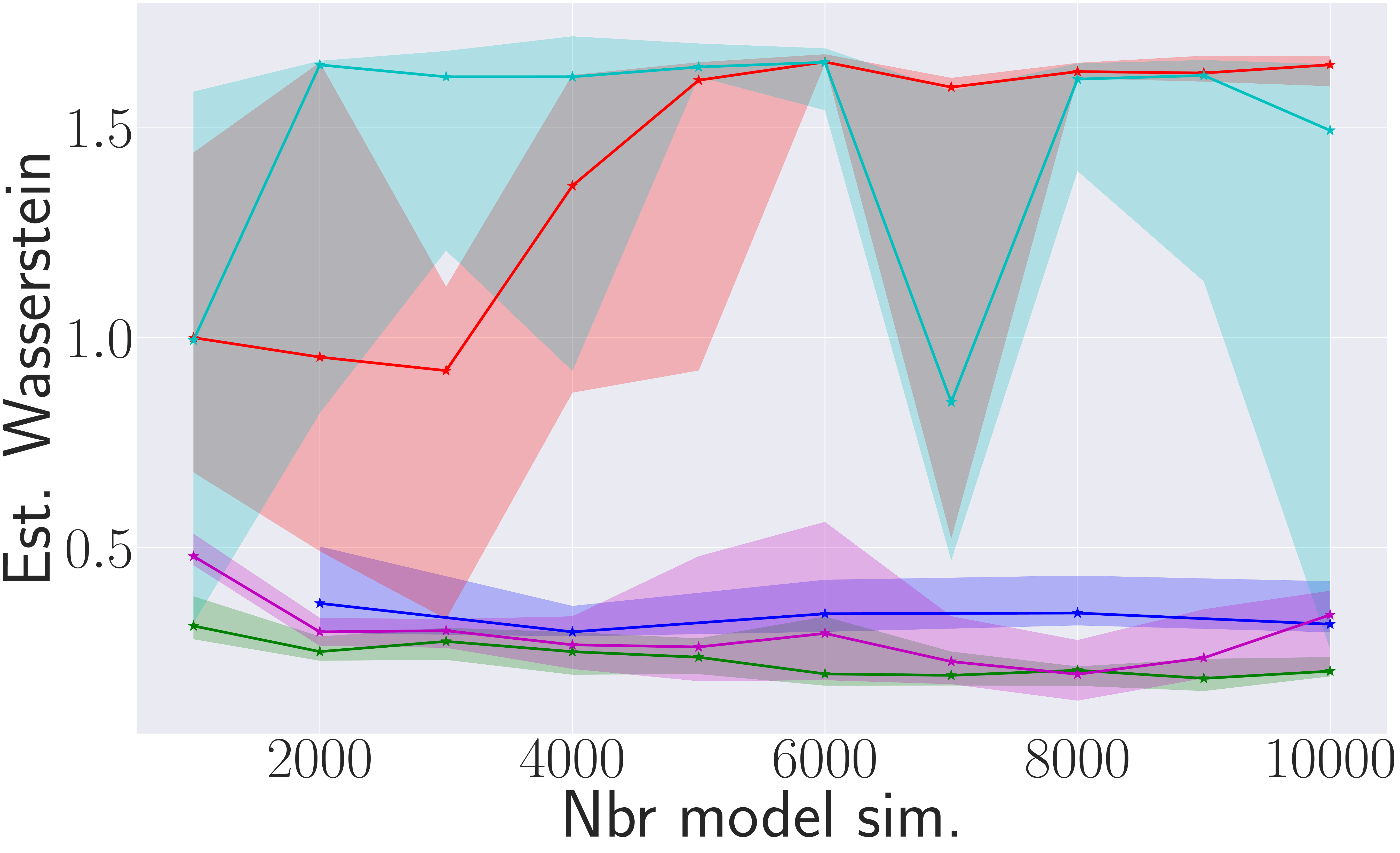} \label{fig:TM_wasser}}
\subfigure[Lotka-Volterra]{\includegraphics[width=.3\columnwidth]{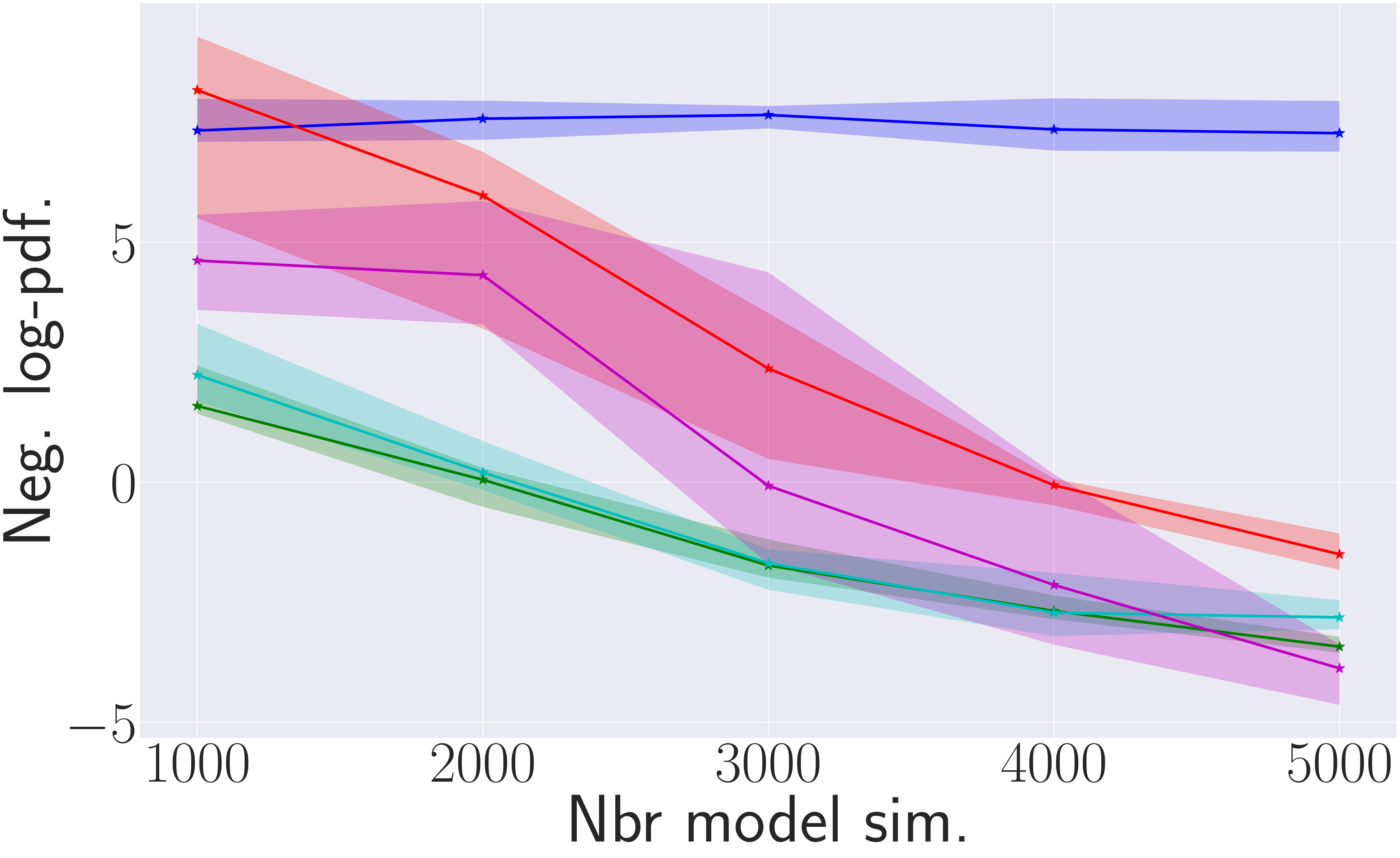} \label{fig:LV_neg_pdf}}
\subfigure[Hodgkin-Huxley]{\includegraphics[width=.3\columnwidth]{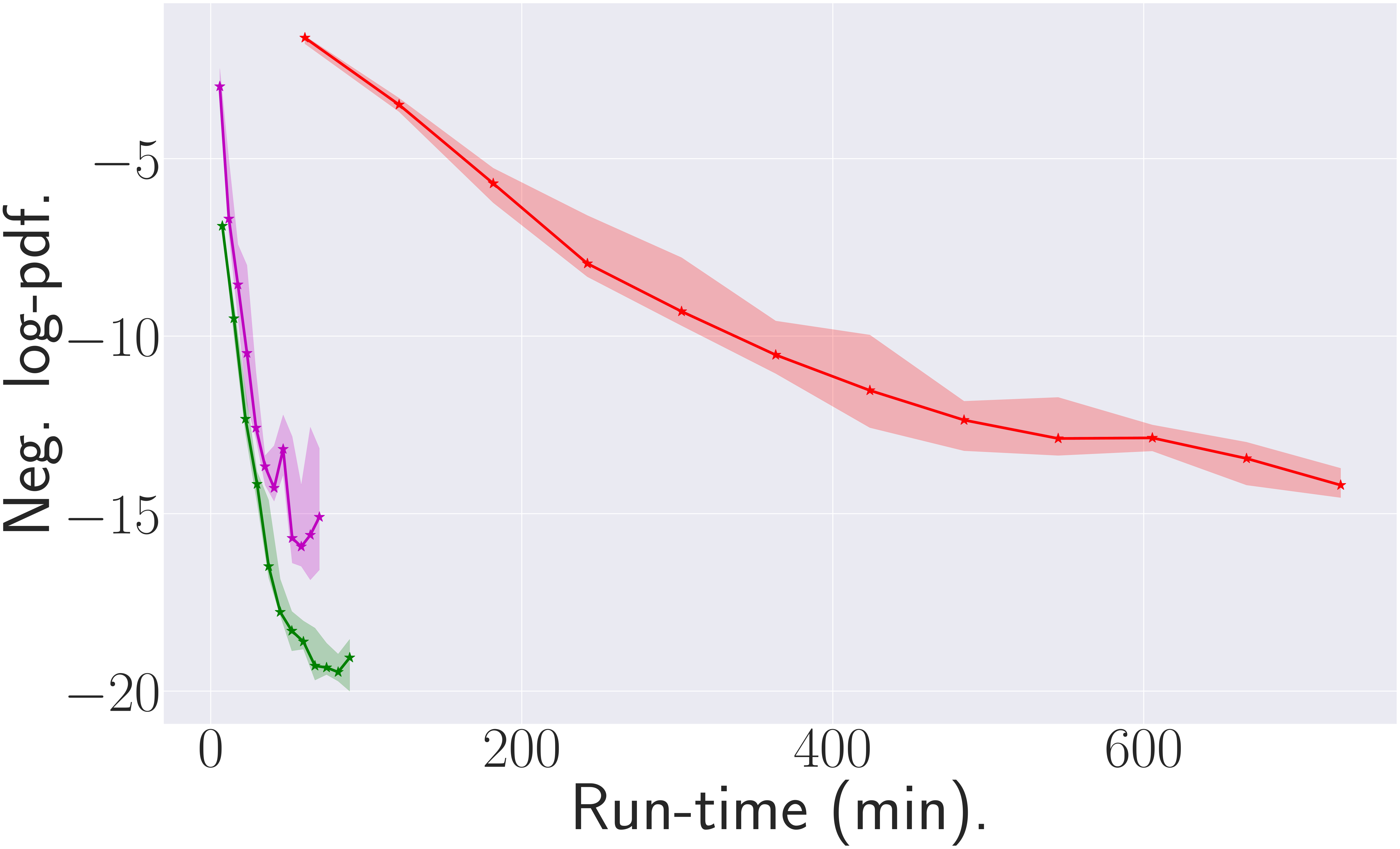} \label{fig:HH_neg_pdf}}
\caption{MV Gaussian/Two-moons/Lotka-Volterra/Hodgkin-Huxley: Posterior inference results for ``five observations'', ``summary statistics'', ``learnable summary statistics'', Two-moons, Lotka-Volterra, and Hodgkin-Huxley. SMC-ABC results (blue); SNL (red); SNPE-C (green); SNRE-B (cyan); SNPLA (magenta). The solid lines show the median value over the attempts and the shaded areas show the range for the 25th and 75th percentile.} 
\label{fig:mv_gauss_post_inferenc_res}
\end{center}
\end{figure}

Posterior inference for the three versions of the MV Gaussian model is in Figures \ref{fig:mv_gauss_post_inferenc_res_five}, \ref{fig:mv_gauss_post_inferenc_res_ss}, and \ref{fig:mv_gauss_post_inferenc_res_ls}  (samples from the resulting posteriors are presented as supplementary material). We conclude that all methods, except SMC-ABC, perform similarly well for ``five observations''. For ``summary statistics'' SNPLA performs the best, followed by SNPE-C and SNL, while both SMC-ABC and SNRE-B under-perform for the given numbers of model simulations. For ``learnable summary statistics'' we have that SNPLA and SNRE-B are converging slightly worse compared to SNL, SNPE-C. 

For ``summary statistics'' we also check the performance of the likelihood model in SNL and SNPLA by sampling from the approximate posterior predictive distribution obtained from the learned likelihood models. 
That is, we sampled summary statistics for $1{,}000$ times from the trained likelihood model $\tilde{p}_{\phi_L}(x|\theta)$. These results are presented in Figure \ref{fig:mv_gauss_post_pred}, and we conclude that SNL and SNPLA perform similarly well. The samples from the approximate posterior predictive distributions also match well with the samples from the analytical posterior predictive. 

\subsection{Complex posterior: Two-moons}

The two-moons  (TM) example \citep{greenberg19aautopost} is a more complex static model. An interesting feature is that the posterior, in some cases, is crescent-shaped. For a technical description of the model, see the supplementary material. 
We ran all methods 10 times (each time with the same observed data set) for $R = 10$ iterations and using $N = 1{,}000$ model simulations for each iteration, and for SNPLA $N_P = 60{,}000$ samples. However, for SMC-ABC we instead used $R = 5$ with $N=2{,}000$. For this experiment we cannot use the KL divergence to evaluate the performance of the posterior inference, and we therefore evaluate the posterior accuracy via the Wasserstein distance between the analytical posterior and the approximate posteriors (for details see the supplementary material).

Inference results are in Figures \ref{fig:TM_wasser} and \ref{fig:TM_post_samples}. We conclude that SNPE-C performs the best. SNPLA also performs well, in particular in many cases the distances from SNPLA are smaller than those from SMC-ABC, however the variability in performance for SNPLA is larger. Indeed while SMC-ABC manages to ``visit'' both crescent moons, the extent of the exploration for each moon is quite poor (notice in Figure \ref{fig:TM_post_samples} it is only apparent that SMC-ABC has fewer samples than claimed; actually many of these overlap on top of each-other). 
Finally, SNL and SNRE-B perform significantly worse than the other methods. SNL and SNRE-B presumably do not perform well here since both use an MCMC sampler, which can struggle to efficiently explore the bimodal target. These results are in line with those in \citet{greenberg19aautopost}.  
We also compare the approximate likelihood models that we learned via SNL and SNPLA. To this end, we simulated data from the learned likelihood models at parameter values simulated from the true posterior. To investigate the performance of the likelihood models we compute the estimated Wasserstein distance between samples from the analytical likelihood and the likelihood models of SNL and SNPLA. The median and quantiles ($Q_{25}, Q_{75}$) of the distances follow: SNL:  $0.111, \,\, (0.103, \,\,  0.136)$,  SNPLA: $0.146, \,\, (0.121, \,\, 0.164)$. Thus, we again conclude that the likelihood models from SNL and SNPLA perform similarly. 

\subsection{Time-series with summary statistics: Lotka-Volterra}

Here, we consider the Lotka-Volterra (LV) case study from \citet{papamakarios19snl}. Observations are assumed to be a set of 9 summary statistics computed from the 2-dimensional time-series (for details, see the supplementary material). We ran all methods 10 times (each time with the same observed data set) for $R=5$ iterations each using $N=1{,}000$ model simulations, and for SNPLA $N_P = 10{,}000$ samples. We followed \citet{papamakarios19snl} and evaluated the obtained posterior distribution by computing its negative log-pdf at the ground-truth parameters (for details, see the supplementary material).   

Results are in Figure \ref{fig:LV_neg_pdf} (samples from the resulting posterior distributions are presented in the supplementary material). For a large number of model simulations, SNPE-C, SNRE-B, and SNPLA perform similarly well. However, for $N<4{,}000$ SNPLA is less precise than SNPE-C and SNRE-B.
The true posterior is not known for this experiment and we therefore evaluate the quality of the posterior by utilizing the simulation-based calibration (SBC) procedure \cite{talts2018validating}. The SBC results for the first parameter are presented Figure \ref{fig:sbc_lv} (the results for the other parameters are presented in the supplementary material). The SBC results show that SNPLA and SNL produce posteriors that are not as well calibrated as the ones obtained from SNPE-C and SNRE-B. The SBC results for SNL are inline with the results presented in \cite{papamakarios19snl}. It could be the case that SNPLA produced worse calibrated posterior results due to having a more complex task to perform.    

We also ran posterior predictive simulations (results presented in the supplementary material), that is we simulated from the \textit{true} model conditionally on the obtained parameter posterior draws. We conclude that posterior predictive simulations from SMC-ABC and SNL do not correspond well with the observed data. However, for SNPE-C, SNPR-B, and SNPLA we observe a more realistic behaviour. The high variability in the model simulations for all methods is due to the intrinsic stochasticity of the LV model which is characteristic of the model even when supplying realistic parameter values.

\begin{figure}[tb]
\begin{center}
\subfigure[LV]{\includegraphics[width=.45\columnwidth]{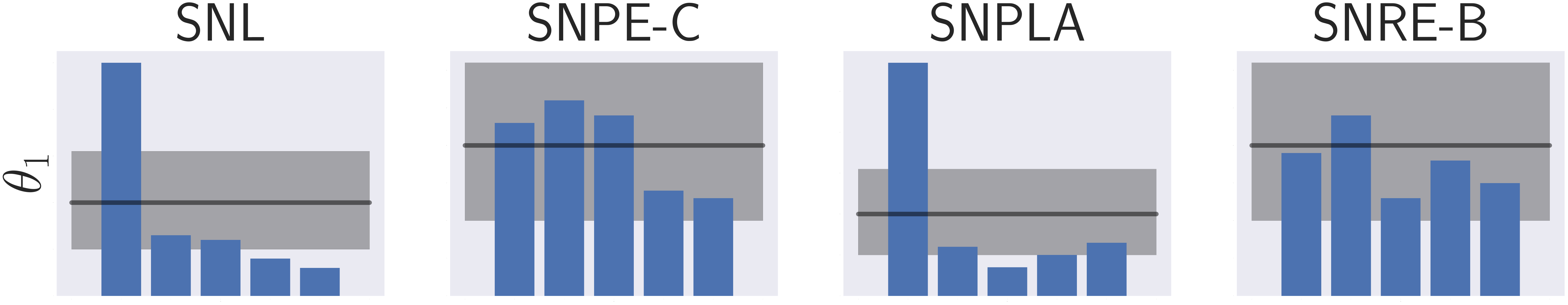} \label{fig:sbc_lv}}
\subfigure[HH]{\includegraphics[width=.45\columnwidth]{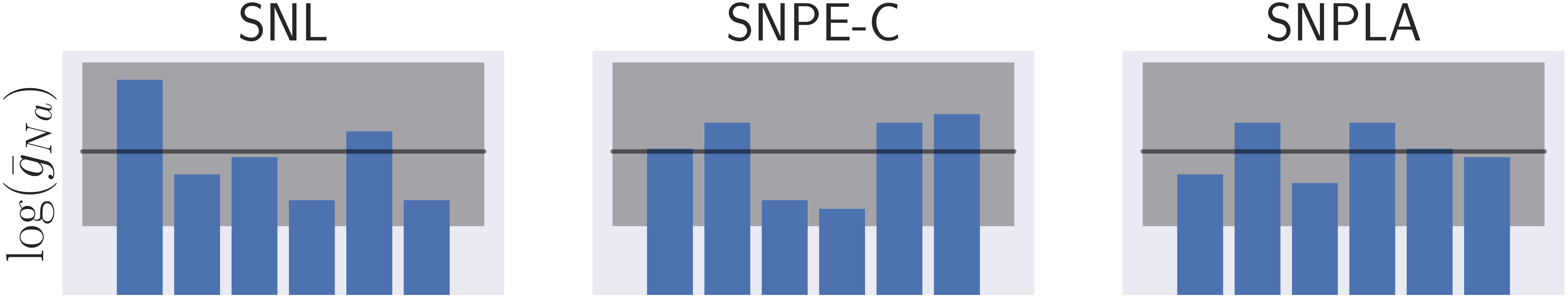} \label{fig:sbc_hh}}
\caption{LV and HH: simulation-based calibration for (a) LV (for the first parameter i.e. $\theta_1$) and (b) HH (for the first parameter i.e. $\log(\bar{g}_{Na})$). Histogram's counts falling within the grey areas denote good calibration.} 
\end{center}
\end{figure}

\subsection{Neural model: Hodgkin-Huxley model}\label{sec:HH}

We now consider the Hodgkin-Huxley (HH) model \cite{hodgkin1952quantitative}, which is used in neuroscience to model the dynamics of a neuron's membrane potential as a function of some stimulus (injected current) and a set of parameters. Following \cite{lueckmann2017flexible, papamakarios19snl}, the likelihood $p(x|\theta)$ is defined on 19 summary statistics computed from simulated voltage time-series. Further details about the model can be found in the supplementary material.
For our simulation-study, we have 10 unknown parameters, and the observed data was simulated from the model.

We ran SNPE-C, SNL and SNPLA 10 times for $R=12$ iterations with $N=2{,}000$ model simulations for each iterations. For SNPLA, we used $N_P = 10{,}000$ samples for each iteration. For each method, the quality of the posterior inference was evaluated by computing the corresponding (approximate) negative log-pdf of the posterior at the ground-truth parameter. We also investigate the quality of the posterior approximation by utilizing SBC analysis, and running posterior predictive simulations. The posterior predictive simulations are obtained by sampling parameter values from the resulting posterior approximation, and then running the HH model simulator at these parameter values.

Posterior inference from using the same number of model simulations is in Figure \ref{fig:HH_neg_pdf}. From Figure \ref{fig:HH_neg_pdf} we note that, in terms of the negative log-pdf of the posterior evaluated at the ground truth parameters and for the same number of model simulations, SNL and SNPLA performs similarly well, however,  SNPLA converges much faster than SNL in terms of wall-clock time. For instance, obtaining a negative log-pdf value of $-15$ takes $69$ minutes for SNPLA, while SNL obtains the same accuracy in $726$ minutes. The SBC results for the first parameter in Figure \ref{fig:sbc_hh} however indicate that all methods produce well-calibrated posteriors (the SBC results for the other parameters are provided in the supplementary material). 
Samples from the resulting posterior approximations are presented in the supplementary material, and these show that SNPE-C and SNPLA perform better than SNL. Also, the approximate marginals overall resemble the inference results presented in \citet{papamakarios19snl} and \citet{lueckmann2017flexible}, with the exception of the $\sigma$ parameter, which could be due to the somewhat different experimental setting we use. Posterior predictive simulations (presented in the supplementary material) show that SNPE-C and SNPLA perform the same and, to a greater extent, resemble the observed data, while SNL is somewhat slightly worse.


To check the quality of the learned likelihood model, we investigate if we can use the latter to rapidly scan for parameter proposals generating data that are similar to the observed data. This is achieved by sampling parameter values from the prior and then computing the number of spikes in the associated simulated data (the number of spikes is one of the summary statistics in the learned likelihood), both when using the true model and when using the trained likelihood model. We then scan the proposals that have produced 4-8 spikes (i.e. parameters that generated data similar to the observed data). For this analysis, the likelihood model agrees with the true model in $~75 \%$ of the times. Scanning 1000 proposals via the true likelihood model took $4500$ sec., while the learned likelihood model utilized $0.2$ sec. for the same task, a $2.3\times 10^4$ acceleration in the runtime. 

\subsection{Hyper-parameter sensitivity analysis} \label{sec:sens_analysis}

The left sub-table of Table \ref{tab:res_hyp_param} reports the results from an analysis where we run each method for ten times, each time with randomly selected hyper-parameters, and always using the same data. We conclude that SNL and SNPLA are the methods that are the most sensitive to the hyper-parameters. 

We also ran SNPLA for 10 different $\lambda$ values (with $\lambda$ = 0.6, \ldots, 0.95), on the same data (results presented in the supplementary material), for all experiments and keeping all other hyper-parameters fixed. This analysis shows that the rate of convergence depends on the value of $\lambda$. However, SNPLA produced a reasonable approximation of the posterior for all attempted $\lambda$'s. 

\begin{table}[tb]
\caption{Left sub-table: Robust coefficient of variation $\frac{IQR}{\text{median}} \cdot 0.75$ (the closer to zero the better) of the  performance measures. Right sub-table: Median run-time (in sec.) for generating $1{,}000$ samples from the resulting posterior. MV Gaussian cases: (i) is ``five observations'', (ii) is ``summary statistics'', and (iii) is ``learnable summary statistics''.} 
\label{tab:res_hyp_param}
\centering
\begin{tabular}{lcccccccc} \toprule
& \multicolumn{4}{c}{Robust coefficient of variation} & \multicolumn{4}{c}{Runtime (sec.)} \\ \cmidrule(r){2-5} \cmidrule(r){6-9}
 Experiment & SNL & SNPE-C & SNPLA & SNRE-B & SNL & SNPE-C & SNPLA & SNRE-B  \\ \midrule
MV-G (i)  &   0.607  &  0.550       &    {0.689}  &    0.421  & 290  &  0.028     & 0.027  & 96  \\   
MV-G (ii) & {766}  &  0.473       &  0.851     & 0.926 & 1{,}841 & 0.024 & 0.042 & 118 \\ 
MV-G (iii)  & {0.649}        &   0.345    &   0.547   & 0.421 &   287   & 0.027    & 0.046 & 88   \\
LV  &  {-4.739}         &  -0.192     &  -0.263    &  -0.249  &   2{,}294  &   0.083   & 0.085    & 166 \\ 
TM    & 0.008         &   0.223     &  {1.108}  &   0.007 &  303     &  0.028 & 0.023  & 77 \\ 
HH    & NA         &   NA     &   NA &   NA &   1{,}824  &   0.167   & 0.197    & NA  \\ \bottomrule 
\end{tabular}
\end{table}

\subsection{Run-time analysis} \label{sec:run_time_analysis}
The run-times for generating $1{,}000$ posterior draws are in the right sub-table  of Table \ref{tab:res_hyp_param}. We conclude that SNPLA on average generates posterior samples $12{,}000$ times faster than SNL. Thus, when the number of model simulations are the same we have that SNPLA will train faster than SNL and SNRE-B In our experiments all methods have access to the same number of model simulations and we have that the training run-time for SNPLA is on average 5.6 times faster compared to SNL, and 2 times faster compared to SNRE-B (the training run-times are reported in the supplementary material). 

\section{Discussion} \label{discussion}
In four case studies, we have shown that SNPLA produces similar posterior inference as other simulation-based algorithms when all methods have access to the same number of model simulations. This is a rather interesting finding since the learning task for SNPLA is more complex compared to the other methods that we compare with, given that SNPLA is set to learn both the posterior model and the likelihood model. However, the variability of the inference obtained with SNPLA is  somewhat consistently higher, which suggests that SNPLA would need to access more model simulations to obtain inference results \textit{at par} with SNL and SNPE-C. Considering SNPLA's complex learning task that would, however, not be surprising.

The computational acceleration in posterior sampling brought by using normalizing flows modelling is staggering. SNPLA (and SNPE-C) generates posterior samples thousands of times more efficiently than SNL and SNRE-B, since draws from SNPLA and SNPE-C are generated by a forward pass of a normalizing flow network, while those from SNL and SNRE-B are generated via MCMC. In Section \ref{sec:run_time_analysis} we have shown that this leads to a substantial speed-up in terms of 
training run-time for SNPLA compared to SNL and SNRE-B. For the HH model, we have also shown that SNPLA can converge faster than SNL in terms of wall-clock time.


Recently, for simulation-based methods, it has been discussed  \citep{durkan2018sequential} if it is of advantage to learn the likelihood or the posterior. With this work we show that this question can be circumvented by learning both. 
However, for posterior inference it seems beneficial to use SNPE-C over SNPLA (at least for the considered examples), but if it is of interest to learn simultaneously the parameters and a cheap model simulator then SNPLA offers this possibility, unlike other considered methods. 
An advantage with learning both is that we obtain an approximation of the distribution that is typically of interest, i.e.~the posterior, and also obtain an approximate model simulator via the learned likelihood model. Since the latter is a normalizing flow, this opens the possibility for the rapid simulation of artificial data, when the ``true simulator'' $p(x|\theta)$ cannot be used more than a handful of times due to e.g. computational constraints. For the HH model we show that the approximate likelihood model can be used for-instance to rapidly scan parameter proposals from the prior predictive. To achieve this, we do not require any semi-supervised learning (human-intervention based labelling) unlike in \cite{wrede2019smart}.  In other contexts, we could generate many samples from the approximate likelihood to estimate tail probabilities for rare events via (otherwise expensive) Monte Carlo simulations. Of course the uncertainty in the approximate simulator's output will be larger when imputed parameters are very different from those learned from the actual data.

\paragraph{Ethics} We have presented an algorithm that learns the posterior and likelihood function of an implicit model. 
However, implicit models could be used for malicious intents, and our algorithm can also be used for such applications.

\section*{Acknowledgements}
The computations were enabled by resources provided by the Swedish National Infrastructure for Computing (SNIC) at LUNARC partially funded by the Swedish Research Council through grant agreement no. 2018-05973. We would also like to thank the developers of the following \texttt{Python} packages for providing the software tools used in this paper: \texttt{sbi} \citep{tejero-cantero2020sbi}, \texttt{nflows} \citep{nflows}, \texttt{PyTorch} \citep{paszke2019torch}, and \texttt{POT: Python Optimal Transport} \citep{flamary2017pot}. UP acknowledges support from the Swedish Research Council (Vetenskapsrådet 2019-03924) and the Chalmers AI Research Centre (CHAIR).

\bibliography{paper}

\begin{thebibliography}{35}
\providecommand{\natexlab}[1]{#1}
\providecommand{\url}[1]{\texttt{#1}}
\expandafter\ifx\csname urlstyle\endcsname\relax
  \providecommand{\doi}[1]{doi: #1}\else
  \providecommand{\doi}{doi: \begingroup \urlstyle{rm}\Url}\fi

\bibitem[Andrieu and Roberts(2009)]{andrieu2009pseudo}
C.~Andrieu and G.~O. Roberts.
\newblock The pseudo-marginal approach for efficient {M}onte {C}arlo
  computations.
\newblock \emph{The Annals of Statistics}, 37:\penalty0 697--725, 2009.

\bibitem[Andrieu et~al.(2010)Andrieu, Doucet, and
  Holenstein]{andrieu2010particle}
C.~Andrieu, A.~Doucet, and R.~Holenstein.
\newblock Particle {M}arkov chain {M}onte {C}arlo methods.
\newblock \emph{Journal of the Royal Statistical Society: Series B (Statistical
  Methodology)}, 72\penalty0 (3):\penalty0 269--342, 2010.

\bibitem[Beaumont et~al.(2002)Beaumont, Zhang, and
  Balding]{beaumont2002approximate}
M.~A. Beaumont, W.~Zhang, and D.~J. Balding.
\newblock Approximate bayesian computation in population genetics.
\newblock \emph{Genetics}, 162\penalty0 (4):\penalty0 2025--2035, 2002.

\bibitem[Beaumont et~al.(2009)Beaumont, Cornuet, Marin, and
  Robert]{beaumont2009adaptive}
M.~A. Beaumont, J.-M. Cornuet, J.-M. Marin, and C.~P. Robert.
\newblock Adaptive approximate bayesian computation.
\newblock \emph{Biometrika}, 96\penalty0 (4):\penalty0 983--990, 2009.

\bibitem[Carnevale and Hines(2006)]{carnevale2006neuron}
N.~T. Carnevale and M.~L. Hines.
\newblock \emph{The NEURON book}.
\newblock Cambridge University Press, 2006.

\bibitem[Cranmer et~al.(2020)Cranmer, Brehmer, and Louppe]{cranmer2019frontier}
K.~Cranmer, J.~Brehmer, and G.~Louppe.
\newblock The frontier of simulation-based inference.
\newblock \emph{Proceedings of the National Academy of Sciences}, 2020.
\newblock \doi{10.1073/pnas.1912789117}.

\bibitem[Dinh et~al.(2017)Dinh, Sohl-Dickstein, and Bengio]{dinh2017rnvp}
L.~Dinh, J.~Sohl-Dickstein, and S.~Bengio.
\newblock Density estimation using real nvp.
\newblock 2017.
\newblock URL \url{https://arxiv.org/abs/1605.08803}.

\bibitem[Durkan et~al.(2018)Durkan, Papamakarios, and
  Murray]{durkan2018sequential}
C.~Durkan, G.~Papamakarios, and I.~Murray.
\newblock Sequential neural methods for likelihood-free inference.
\newblock \emph{arXiv preprint arXiv:1811.08723}, 2018.

\bibitem[Durkan et~al.(2019)Durkan, Bekasov, Murray, and
  Papamakarios]{durkan2019neural}
C.~Durkan, A.~Bekasov, I.~Murray, and G.~Papamakarios.
\newblock Neural spline flows.
\newblock In \emph{Advances in Neural Information Processing Systems}, pages
  7511--7522, 2019.

\bibitem[Durkan et~al.(2020{\natexlab{a}})Durkan, Bekasov, Murray, and
  Papamakarios]{nflows}
C.~Durkan, A.~Bekasov, I.~Murray, and G.~Papamakarios.
\newblock {nflows}: normalizing flows in {PyTorch}, Nov. 2020{\natexlab{a}}.
\newblock URL \url{https://doi.org/10.5281/zenodo.4296287}.

\bibitem[Durkan et~al.(2020{\natexlab{b}})Durkan, Murray, and
  Papamakarios]{durkan2020contrastive}
C.~Durkan, I.~Murray, and G.~Papamakarios.
\newblock On contrastive learning for likelihood-free inference.
\newblock In H.~D. III and A.~Singh, editors, \emph{Proceedings of the 37th
  International Conference on Machine Learning}, volume 119 of
  \emph{Proceedings of Machine Learning Research}, pages 2771--2781,
  2020{\natexlab{b}}.

\bibitem[Flamary and Courty(2017)]{flamary2017pot}
R.~Flamary and N.~Courty.
\newblock Pot python optimal transport library, 2017.
\newblock URL \url{https://pythonot.github.io/}.

\bibitem[Greenberg et~al.(2019)Greenberg, Nonnenmacher, and
  Macke]{greenberg19aautopost}
D.~Greenberg, M.~Nonnenmacher, and J.~Macke.
\newblock Automatic posterior transformation for likelihood-free inference.
\newblock In K.~Chaudhuri and R.~Salakhutdinov, editors, \emph{Proceedings of
  the 36th International Conference on Machine Learning}, volume~97 of
  \emph{Proceedings of Machine Learning Research}, pages 2404--2414, 2019.

\bibitem[Gutmann and Corander(2016)]{gutmann2016bayesian}
M.~U. Gutmann and J.~Corander.
\newblock Bayesian optimization for likelihood-free inference of
  simulator-based statistical models.
\newblock \emph{The Journal of Machine Learning Research}, 17\penalty0
  (1):\penalty0 4256--4302, 2016.

\bibitem[Hermans et~al.(2020)Hermans, Begy, and Louppe]{hermans2019likelihood}
J.~Hermans, V.~Begy, and G.~Louppe.
\newblock Likelihood-free {MCMC} with amortized approximate ratio estimators.
\newblock In H.~D. III and A.~Singh, editors, \emph{Proceedings of the 37th
  International Conference on Machine Learning}, volume 119 of
  \emph{Proceedings of Machine Learning Research}, pages 4239--4248, 2020.

\bibitem[Hodgkin and Huxley(1952)]{hodgkin1952quantitative}
A.~L. Hodgkin and A.~F. Huxley.
\newblock A quantitative description of membrane current and its application to
  conduction and excitation in nerve.
\newblock \emph{The Journal of physiology}, 117\penalty0 (4):\penalty0
  500--544, 1952.

\bibitem[Kobyzev et~al.(2020)Kobyzev, Prince, and
  Brubaker]{kobyzev2020normalizing}
I.~Kobyzev, S.~Prince, and M.~Brubaker.
\newblock Normalizing flows: An introduction and review of current methods.
\newblock \emph{IEEE Transactions on Pattern Analysis and Machine
  Intelligence}, 2020.

\bibitem[Lueckmann et~al.(2017)Lueckmann, Goncalves, Bassetto, {\"O}cal,
  Nonnenmacher, and Macke]{lueckmann2017flexible}
J.-M. Lueckmann, P.~J. Goncalves, G.~Bassetto, K.~{\"O}cal, M.~Nonnenmacher,
  and J.~H. Macke.
\newblock Flexible statistical inference for mechanistic models of neural
  dynamics.
\newblock In \emph{Advances in Neural Information Processing Systems}, pages
  1289--1299, 2017.

\bibitem[Lueckmann et~al.(2021)Lueckmann, Boelts, Greenberg, Gon{\c{c}}alves,
  and Macke]{lueckmann2101benchmarking}
J.-M. Lueckmann, J.~Boelts, D.~S. Greenberg, P.~J. Gon{\c{c}}alves, and J.~H.
  Macke.
\newblock Benchmarking simulation-based inference.
\newblock \emph{arXiv preprint arXiv:2101.04653}, 2021.

\bibitem[Marin et~al.(2012)Marin, Pudlo, Robert, and
  Ryder]{marin2012approximate}
J.-M. Marin, P.~Pudlo, C.~P. Robert, and R.~J. Ryder.
\newblock Approximate bayesian computational methods.
\newblock \emph{Statistics and Computing}, 22\penalty0 (6):\penalty0
  1167--1180, 2012.

\bibitem[Papamakarios and Murray(2016)]{papamakarios2016fast}
G.~Papamakarios and I.~Murray.
\newblock Fast $\varepsilon$-free inference of simulation models with bayesian
  conditional density estimation.
\newblock In \emph{Advances in Neural Information Processing Systems}, pages
  1028--1036, 2016.

\bibitem[Papamakarios et~al.(2017)Papamakarios, Pavlakou, and
  Murray]{papamakarios2017masked}
G.~Papamakarios, T.~Pavlakou, and I.~Murray.
\newblock Masked autoregressive flow for density estimation.
\newblock In \emph{Advances in Neural Information Processing Systems}, pages
  2338--2347, 2017.

\bibitem[Papamakarios et~al.(2019{\natexlab{a}})Papamakarios, Nalisnick,
  Rezende, Mohamed, and Lakshminarayanan]{papamakarios2019normalizing}
G.~Papamakarios, E.~Nalisnick, D.~J. Rezende, S.~Mohamed, and
  B.~Lakshminarayanan.
\newblock Normalizing flows for probabilistic modeling and inference.
\newblock \emph{arXiv preprint arXiv:1912.02762}, 2019{\natexlab{a}}.

\bibitem[Papamakarios et~al.(2019{\natexlab{b}})Papamakarios, Sterratt, and
  Murray]{papamakarios19snl}
G.~Papamakarios, D.~Sterratt, and I.~Murray.
\newblock Sequential neural likelihood: Fast likelihood-free inference with
  autoregressive flows.
\newblock In K.~Chaudhuri and M.~Sugiyama, editors, \emph{Proceedings of
  Machine Learning Research}, volume~89 of \emph{Proceedings of Machine
  Learning Research}, pages 837--848. PMLR, 16--18 Apr 2019{\natexlab{b}}.
\newblock URL \url{http://proceedings.mlr.press/v89/papamakarios19a.html}.

\bibitem[Paszke et~al.(2019)Paszke, Gross, Massa, Lerer, Bradbury, Chanan,
  Killeen, Lin, Gimelshein, Antiga, Desmaison, Kopf, Yang, DeVito, Raison,
  Tejani, Chilamkurthy, Steiner, Fang, Bai, and Chintala]{paszke2019torch}
A.~Paszke, S.~Gross, F.~Massa, A.~Lerer, J.~Bradbury, G.~Chanan, T.~Killeen,
  Z.~Lin, N.~Gimelshein, L.~Antiga, A.~Desmaison, A.~Kopf, E.~Yang, Z.~DeVito,
  M.~Raison, A.~Tejani, S.~Chilamkurthy, B.~Steiner, L.~Fang, J.~Bai, and
  S.~Chintala.
\newblock Pytorch: An imperative style, high-performance deep learning library.
\newblock In H.~Wallach, H.~Larochelle, A.~Beygelzimer, F.~d\textquotesingle
  Alch\'{e}-Buc, E.~Fox, and R.~Garnett, editors, \emph{Advances in Neural
  Information Processing Systems 32}, pages 8024--8035. Curran Associates,
  Inc., 2019.
\newblock URL
  \url{http://papers.neurips.cc/paper/9015-pytorch-an-imperative-style-high-performance-deep-learning-library.pdf}.

\bibitem[Price et~al.(2018)Price, Drovandi, Lee, and Nott]{price2018bayesian}
L.~F. Price, C.~C. Drovandi, A.~Lee, and D.~J. Nott.
\newblock Bayesian synthetic likelihood.
\newblock \emph{Journal of Computational and Graphical Statistics}, 27\penalty0
  (1):\penalty0 1--11, 2018.

\bibitem[Radev et~al.(2020)Radev, Mertens, Voss, Ardizzone, and
  K{\"o}the]{radev2020bayesflow}
S.~T. Radev, U.~K. Mertens, A.~Voss, L.~Ardizzone, and U.~K{\"o}the.
\newblock Bayesflow: Learning complex stochastic models with invertible neural
  networks.
\newblock \emph{IEEE Transactions on Neural Networks and Learning Systems},
  2020.

\bibitem[Rezende and Mohamed(2015)]{rezende2015variational}
D.~Rezende and S.~Mohamed.
\newblock Variational inference with normalizing flows.
\newblock In \emph{International Conference on Machine Learning}, pages
  1530--1538. PMLR, 2015.

\bibitem[Talts et~al.(2018)Talts, Betancourt, Simpson, Vehtari, and
  Gelman]{talts2018validating}
S.~Talts, M.~Betancourt, D.~Simpson, A.~Vehtari, and A.~Gelman.
\newblock Validating bayesian inference algorithms with simulation-based
  calibration.
\newblock \emph{arXiv preprint arXiv:1804.06788}, 2018.

\bibitem[Tejero-Cantero et~al.(2020)Tejero-Cantero, Boelts, Deistler,
  Lueckmann, Durkan, Gonçalves, Greenberg, and Macke]{tejero-cantero2020sbi}
A.~Tejero-Cantero, J.~Boelts, M.~Deistler, J.-M. Lueckmann, C.~Durkan, P.~J.
  Gonçalves, D.~S. Greenberg, and J.~H. Macke.
\newblock sbi: A toolkit for simulation-based inference.
\newblock \emph{Journal of Open Source Software}, 5\penalty0 (52):\penalty0
  2505, 2020.
\newblock \doi{10.21105/joss.02505}.
\newblock URL \url{https://doi.org/10.21105/joss.02505}.

\bibitem[Thomas et~al.(2020)Thomas, Dutta, Corander, Kaski, and
  Gutmann]{thomas2020likelihood}
O.~Thomas, R.~Dutta, J.~Corander, S.~Kaski, and M.~U. Gutmann.
\newblock Likelihood-free inference by ratio estimation.
\newblock \emph{Bayesian Analysis}, 2020.

\bibitem[Wiqvist et~al.(2019)Wiqvist, Mattei, Picchini, and
  Frellsen]{wiqvist2019pen}
S.~Wiqvist, P.-A. Mattei, U.~Picchini, and J.~Frellsen.
\newblock Partially exchangeable networks and architectures for learning
  summary statistics in approximate {B}ayesian computation.
\newblock In \emph{Proceedings of the 36th International Conference on Machine
  Learning}. PMLR, 2019.

\bibitem[Wood(2010)]{wood2010statistical}
S.~N. Wood.
\newblock Statistical inference for noisy nonlinear ecological dynamic systems.
\newblock \emph{Nature}, 466\penalty0 (7310):\penalty0 1102--1104, 2010.

\bibitem[Wrede and Hellander(2019)]{wrede2019smart}
F.~Wrede and A.~Hellander.
\newblock Smart computational exploration of stochastic gene regulatory network
  models using human-in-the-loop semi-supervised learning.
\newblock \emph{Bioinformatics}, 35\penalty0 (24):\penalty0 5199--5206, 2019.

\bibitem[Zaheer et~al.(2017)Zaheer, Kottur, Ravanbakhsh, Poczos, Salakhutdinov,
  and Smola]{zaheer2017deep}
M.~Zaheer, S.~Kottur, S.~Ravanbakhsh, B.~Poczos, R.~R. Salakhutdinov, and A.~J.
  Smola.
\newblock Deep sets.
\newblock In \emph{Advances in Neural Information Processing Systems}, pages
  3391--3401, 2017.

\end{thebibliography}
\bibliographystyle{abbrvnat}


\newpage

\begin{center}
    \section*{Supplementary Material to:\\
Sequential Neural Posterior and Likelihood Approximation} 
\end{center}

\renewcommand*\contentsname{Contents -- supplementary material}

\tableofcontents

\addtocontents{toc}{\protect\setcounter{tocdepth}{6}}


\section{Introduction to the supplementary material}

This document primarily presents additional details on the experimental settings and inference results. However, we also discuss some technicalities on setting up the normalizing flow models.

\section{Normalizing flows for simulation-based inference}
\label{sec:nf_for_simbiasedinference}

Here we introduce some basic notions that are necessary in order to follow our method.
The normalizing flow model, introduced in \citet{rezende2015variational} (see \citet{kobyzev2020normalizing, papamakarios2019normalizing} for reviews), is a probabilistic model that transforms a simple \textit{base distribution} $u \sim p_u(u)$ into some complex distribution $x \sim p_x(x)$ via the following transformation
\begin{equation*}
    x = T(u), \qquad u \sim p_u(u).
\end{equation*}
The function $T$ is parameterized with an \textit{invariant} neural network such that $T^{-1}$ exists, and that both $T$ and $T^{-1}$ are differentiable, i.e.~the function $T$ is \textit{diffeomorphic}. The probability density function (pdf) for $x$ is computed via the Jacobian $J_T$ of $T$ (or $J_{T^{-1}}$ of $T^{-1}$) in the following two equivalent ways:  
\begin{align*}
\begin{cases}
    p(x) &= p_u(u)|\det J_T(u)|^{-1}, \qquad u \sim T^{-1}(x), \\
    p(x) &= p_u(T^{-1}(x))|\det J_{T^{-1}}(x)|.
\end{cases}
\end{align*}
Due to this structure of the pdf, it is easy to construct complex transformations by composing, say, $n$ transformations such that $T = T_1 \circ T_2 \circ \cdot \circ T_n$ where each transformation $T_i$ is diffeomorphic, and where the Jacobian contribution for each $T_i$ can be computed. Flow models that allow for building these kinds of nested structures are, for instance, RNVP \cite{dinh2017rnvp}, Neural Spline Flow \cite{durkan2019neural}, and  Masked Autoregressive Flow \cite{papamakarios2017masked}.  
Now, assume that we have a normalizing flow model $\tilde{p}_x(x; \phi)$ (with weights $\phi$ for neural network $T$), and that our target distribution is denoted $p_x(x)$. We want  to train $\tilde{p}_x(x; \phi)$ so that it approximates  $p_x(x)$. Let us also assume that we can obtain samples from the target $p_x(x)$, then we can use the \textit{forward} KL divergence to fit the flow model  by utilizing the following loss function 
\begin{align}
    \mathcal{L}(\phi) &= D_{KL}\big[p_x(x)  \big|\big| \tilde{p}_x(x; \phi) \big], \nonumber \\
    &= -E_{p_x(x)} \big[ \log \tilde{p}_x(x; \phi) \big] + \text{const.},  \nonumber \\
    &= -E_{p_x(x)} \big[ \log p_u(T^{-1}(x;\theta)) + \log |\det J_{T^{-1}}(x; \phi) \big] + \text{const.}.  \label{eq:loss_forward}
\end{align} 
The loss in \eqref{eq:loss_forward} is typically (and also in our case) evaluated via Monte Carlo. 

If we do not have access to samples from the target distribution $p_x(x)$, but we can evaluate the pdf of $p_x(x)$,  it is possible to fit $\tilde{p}_x(x; \phi)$ via the \textit{reverse} KL divergence using the loss
\begin{align}
    \mathcal{L}(\phi) &= D_{KL}\big[\tilde{p}_x(x; \phi)   \big|\big| p_x(x) \big], \nonumber \\
    &= E_{\tilde{p}_x(x; \phi)} \big[ \log \tilde{p}_x(x; \phi) - \log  p_x(x) \big], \nonumber \\
    &= E_{p_u(u)} \big[ \log p_u(u)) -  \log |\det J_T(u; \phi)| - \log p_x(T(u; \phi)) \big]. \nonumber 
\end{align} 
It is shown in \citet{papamakarios2017masked} that the forward and the reverse KL divergence are equivalent. The possibility to fit the flow model $\tilde{p}_x(x; \beta)$ via the reverse KL divergence is critical for our method, since SNPLA  uses it to train the posterior model without using a proposal distribution.

\section{Pseudo-code for SNL and SNPE}

SNL and SNPE are presented in Algorithms \ref{algo:snl}
and \ref{algo:snpe}. 

\begin{minipage}[t]{0.48\textwidth}
  \vspace{0pt}  
\begin{algorithm}[H]
\caption{SNL}
\label{algo:snl}
\DontPrintSemicolon 
\KwIn{Untrained likelihood model $\tilde{p}_{\phi_L}(x | \theta)$, number of iterations $R$, number of training samples per iteration $N$. }
\KwOut{Trained likelihood model $\tilde{p}_{\phi_L}(x | \theta)$.}
Set $\hat{p}_0(\theta | x_{\text{obs}}) \gets p(\theta)$, $\mathcal{D}=\{\emptyset\}$\;
\For{$r = 1:R$} {
    For $n = 1:N$ sample $$(\theta_n, x_n) \sim  \tilde{p}(\theta,x) = p(x | \theta)\hat{p}_{r-1}(\theta | x_{\text{obs}}).$$

    Update training data $\mathcal{D} = [\theta_{1:N}, x_{1:N}]\cup\mathcal{D}$. \\
    
    Update $\tilde{p}_{\phi_L}(x | \theta)$ by maximizing the following loss
    \begin{align*}
    &\mathcal{L}(\phi_L) = -E_{\tilde{p}(\theta,x)}\big( \log \tilde{p}_{\phi_L}(x | \theta) \big).  
    \end{align*}
    
    Update the proposal distribution, i.e. let  
    $$\hat{p}_{r}(\theta | x_{\text{obs}}) \propto\tilde{p}_{\phi_L}(x_{\text{obs}} | \theta)p(\theta).$$

    \vspace{14pt}
}
\end{algorithm}
\end{minipage}\hspace{0.02\textwidth}
\begin{minipage}[t]{0.48\textwidth}
  \vspace{0pt}
\begin{algorithm}[H]
\caption{SNPE}
\label{algo:snpe}
\DontPrintSemicolon 
\KwIn{Untrained posterior model $\tilde{p}_{\phi_P}(\theta | x)$, number of iterations $R$, number of training samples per iteration $N$.}
\KwOut{Trained posterior model $\tilde{p}_{\phi_P}(\theta | x)$.}
Set $\hat{p}_0(\theta | x_{\text{obs}}) \gets p(\theta)$, $\mathcal{D}=\{\emptyset\}$\;
\For{$r = 1:R$} {
    For $n=1:N$ sample  $$(\theta_n, x_n) \sim  \tilde{p}(\theta,x) = p(x | \theta)\hat{p}_{r-1}(\theta | x_{\text{obs}}).$$
    
    Update training data $\mathcal{D} = [\theta_{1:N}, x_{1:N}]\cup\mathcal{D}$. \\

    Update $\tilde{p}_{\phi_P}(\theta | x)$ by maximizing the following loss
    \begin{align*}
    &\mathcal{L}(\phi_P) = -E_{\tilde{p}(\theta,x)}\big( \log \tilde{p}_{\phi_P}(\theta | x) \big).
    \end{align*}
    
    Update the proposal distribution, i.e. let 
    $$\hat{p}_{r}(\theta | x_{\text{obs}}) \leftarrow \frac{p(\theta)}{\hat{p}_{r-1}(\theta | x_{\text{obs}})}\tilde{p}_{\phi_P}(\theta | x_{\text{obs}}).$$
}
\end{algorithm}
\end{minipage}

\section{Computer environment} \label{sec:comp_env}

The code for replicating the experiments can be found at \href{https://github.com/SamuelWiqvist/snpla}{https://github.com/SamuelWiqvist/snpla}. 
All experiments were implemented in \texttt{Python 3.7.4}, with normalizing flow models built using the \texttt{nflows} package \citep{nflows}. The \texttt{sbi} package \citep{tejero-cantero2020sbi} was used to run SMC-ABC, SNPE-C, SNRE-B, and SNL. We used the \texttt{Neuron} software \citep{carnevale2006neuron} to produce all model simulations for the HH model. Full specifications of the computer environments used are provided in the \texttt{env\_local.ylm} and \texttt{env\_unnamed.txt} files. Regarding licenses for the main already existing assets: \texttt{Python 3.7.4} is licensed under the \textit{PSF License Agreement};  \texttt{sbi} is licensed under the \textit{GNU Affero General Public License v3.0}; \texttt{nflows} is licensed under the \textit{MIT License}; and \texttt{Neuron} is licensed under the \textit{GNU GPL}. For information regarding licenses of the new assets see the \texttt{LICENSE} file.  

\section{Experimental setting}

For all experiments , except the HH model, we compare the following inference methods: SMC-ABC \citep{beaumont2009adaptive}, SNL \citep{papamakarios19snl}, SNPE-C \citep{greenberg19aautopost}, SNRE-B \citep{durkan2020contrastive}, and SNPLA. For HH we only compare SNL, SNPE-C and SNPLA.  
The experiments are set-up so that all methods have access to the same number of model simulations. The normalizing flow models used by SNL, SNPE-C, and SNPLA were set to be the same. The masked autoregressive flow architecture  \cite{papamakarios2017masked} was used to construct the flow models.

\section{Performance measures}

In this section, we give the definitions of the performance measures used. 

\subsection{Kullback–Leibler (KL) divergence}

The KL divergence for two continuous distributions $P$ and $Q$ is defined as 
\begin{align*}
    D_{KL}\big(P | Q\big) = \int p(x)\log \Big( \frac{p(x)}{q(x)} \Big) dx,
\end{align*}
where $p$ and $q$ are the associated probability density functions (pdf). In some cases, the KL divergence is analytically known, and in equation \eqref{eq:kl_div_mv_gauss} below we give the analytical formula for when $P$ and $Q$ are Gaussian. 

\subsection{Wasserstein distance}

The $p^{\text{th}}$ Wasserstein distance between two random variables  $X$ and $Y$ on the metric space $(M,d)$ is given by 
\begin{align*}
    W_p(\mu,\nu) = \big( \inf E[d(X,Y)^p]\big)^{1/p}, 
\end{align*}
where $d$ is the distance function on the associated metric space, $\mu$ and $\nu$ are the marginal distributions of $X$ and $Y$ respectively, and the infimum is taken over all joint distributions of the random variables $X$ and $Y$ with marginals $\mu$ and $\nu$.
For TM we used the \texttt{POT: Python Optimal Transport} package \citep{flamary2017pot} (utilizing the default settings) to estimate the $1^{\text{st}}$ Wasserstein distance.

\subsection{Negative log-pdf at ground-truth parameter}

Assume that we have the posterior approximation $\tilde{p}_{\phi_P}(\theta  | x_{\text{obs}})$, and  that the ground-truth parameter is $\theta^{\star}$. The approximate negative log-pdf of the posterior evaluated at the ground-truth parameter is now given by $\log \tilde{p}_{\phi_P}\theta^{\star}  | x_{\text{obs}})$. In our work we compute the negative log-pdf of ground-truth parameter by first approximating $\tilde{p}_{\phi_P}(\theta  | x_{\text{obs}})$ with a Gaussian distribution, which is estimated from the posterior samples. This step is included to have a consistent approximation of the posterior across models.  

\section{Incorporating uniform priors in the normalizing flow model for the posterior} \label{sec:using_priors_in_post_nf_model}

The normalizing flow posterior model  $\tilde{p}_{\phi_P}(\theta  | x_{\text{obs}})$ learns the composite transformation $T = T_1 \circ T_2 \circ \cdot \circ T_n$ that maps some base distribution $u$ into the parameter posterior distribution $p(\theta  | x_{\text{obs}})$. However, if we have a uniform prior for parameter $\theta \sim U(a,b)$,
then we know that the posterior will only have non-zero mass on the interval $[a,b]$. This information can be leveraged in the normalizing flow posterior model $\tilde{p}_{\phi_P}(\theta  | x_{\text{obs}})$ by adding a scaled and shifted sigmoid function $\sigma_{scale, shift}$ as the last transformation in the chain of transformations that constitutes  $T$. Thus, in that case we have that $T = T_1 \circ T_2 \circ \cdot \circ T_n \circ \sigma_{scale, shift}$. This way, $\tilde{p}_{\phi_P}(\theta  | x_{\text{obs}})$ will have positive mass only on the interval $[a,b]$. This is used in all cases where we have uniform priors.  

\section{Multivariate Gaussian (MV Gaussian)} \label{sec:MV}

\subsection{Model specification}

The conjugate MV Gaussian model of dimension 2 is given by
\begin{align*}
\begin{cases}
    \mu &\sim N(\mu  | \mu_{\mu}, \Sigma_{\mu}), \\ 
    x &\sim N\Bigg(\mu, \Sigma = \begin{bmatrix} 1.3862 & 1.4245 \\ 1.4245 & 1.5986\end{bmatrix}\Bigg).
\end{cases}
\end{align*}
The hyperparameters $\mu_{\mu}$ and $\Sigma_{\mu}$ are set to 
\begin{align*}
    \mu_{\mu} =  \begin{bmatrix} 0 \\ 0\end{bmatrix}, \qquad \Sigma_{\mu} \begin{bmatrix} 5 & 0 \\ 0 & 5\end{bmatrix}. 
\end{align*}
Each simulated dataset was generated by first sampling ground-truth parameters for the mean vector $\mu_{gt}$ from the corresponding prior distribution. We did this for each of the several versions of the MV Gaussian model. 

\subsection{Summary statistics used in the ``summary statistics'' case-study}

We used the following summary statistics for the ``summary statistics'' case: the two sample means of the data, the two sample variances and the sample covariance between the two components, i.e.~the sufficient statistics for the model.

\subsection{Computing the KL divergence}

The KL divergence between the analytical posterior $p(\mu | x, \Sigma)$ and some approximation $p^{\star}(\mu|x)$ is given by 
\begin{align} 
    D_{KL}\big(p(\mu | x, \Sigma\big) \big|\big| p^{\star}(\mu|x)\big) &=  D_{KL}\big(N(\mu | m, \Sigma\big) \big|\big| p^{\star}(\mu|m^{\star}, \Sigma^{\star})\big) \nonumber \\
    &= \frac{1}{2}\Big[ \log \frac{\det\Sigma^{\star^{-1}}}{\det\Sigma^{-1}} + Tr(\Sigma^{\star^{-1}}\Sigma) - D + (m - m^{\star})^{T}\Sigma^{\star^{-1}}(m - m^{\star})\Big], \label{eq:kl_div_mv_gauss}
\end{align}
where $m$ is the mean and $\Sigma$ the covariance matrix of the analytical posterior. Additionally,  $m^{\star}$ and $\Sigma^{\star}$ are the sample mean and sample covariance matrix based on 1,000 samples from the posterior approximation, respectively. $D$ is the dimension of $x$, i.e. $D = 2$ in our case. 

\subsection{Posterior inference}

\begin{figure}[bt]
\begin{center}

\subfigure[``Five observations'' \textcolor{white}{}]{\includegraphics[width=1.\columnwidth]{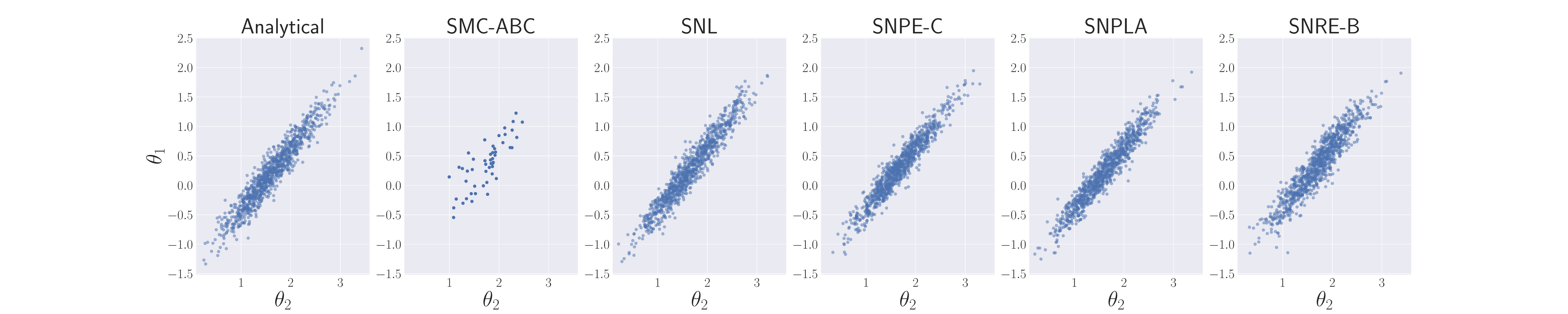}}
\\
\subfigure[``Summary statistics'' \textcolor{white}{}]{\includegraphics[width=1.\columnwidth]{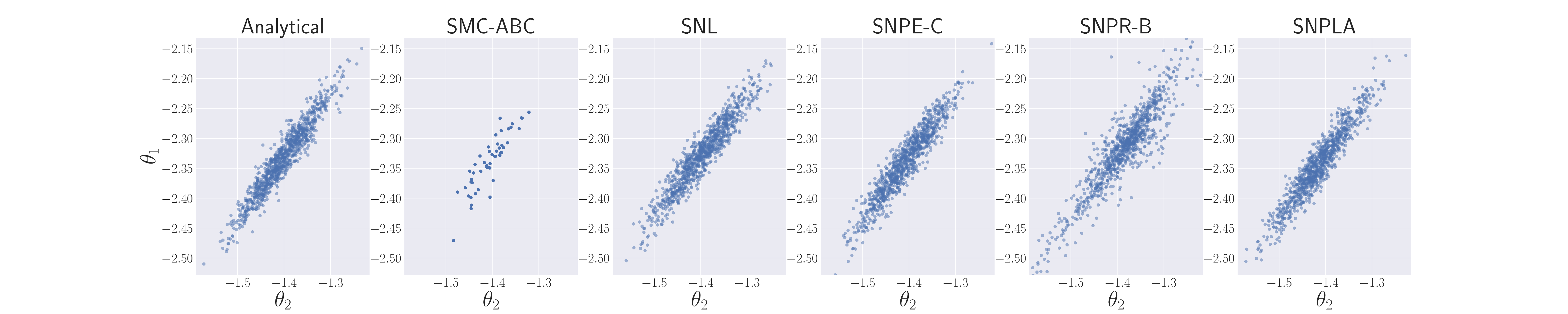}}\qquad
\\
\subfigure[``Learnable summary statistics'' \textcolor{white}{}]{\includegraphics[width=1.\columnwidth]{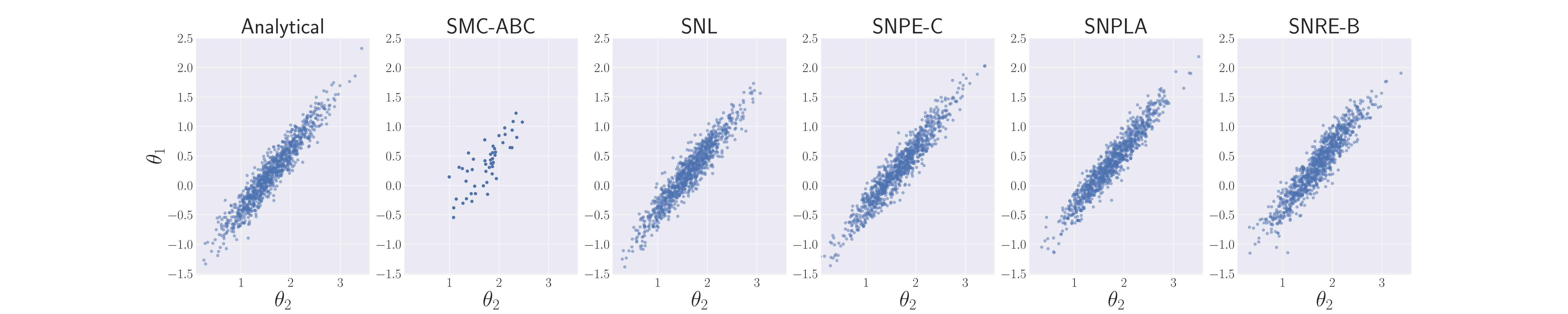}}
\caption{MV:Gaussian: Samples from the resulting posterior approximations for ``five observations'', ``summary statistics'', and ``learnable summary statistics''. Results for one data set.}
\label{fig:mv_gauss_post_samples}
\end{center}
\end{figure}

Samples from the resulting posterior approximations (for one run) are presented in Figure \ref{fig:mv_gauss_post_samples}. For the analysis, see the main paper. 

\section{Complex posterior: Two-moons (TM)} \label{sec:tm}

\subsection{Model specification}

A full technical description for the TM model can be found in the supplementary material for \citet{greenberg19aautopost}. We followed the model specification in \citet{greenberg19aautopost}, thus we set the observed data to be $x_{\text{obs}} = [0,0]^T$.  However, we used a slightly setting of the model and for our experiment we have that: $r \sim N(1.0, 0.1^2)$, and $p = (\cos(a) + 1, r\sin(a))$. Due to this setting we use the following priors for $\theta_i$: 

\begin{align*}
    \theta_i \sim U(-2,2), \qquad i = 1:2. 
\end{align*}

\section{Time-series with summary statistics: Lotka-Volterra (LV)} \label{sec:LV}

\subsection{Model specification}

\begin{figure}[bt]
\begin{center}
\centerline{\includegraphics[width=0.8\columnwidth]{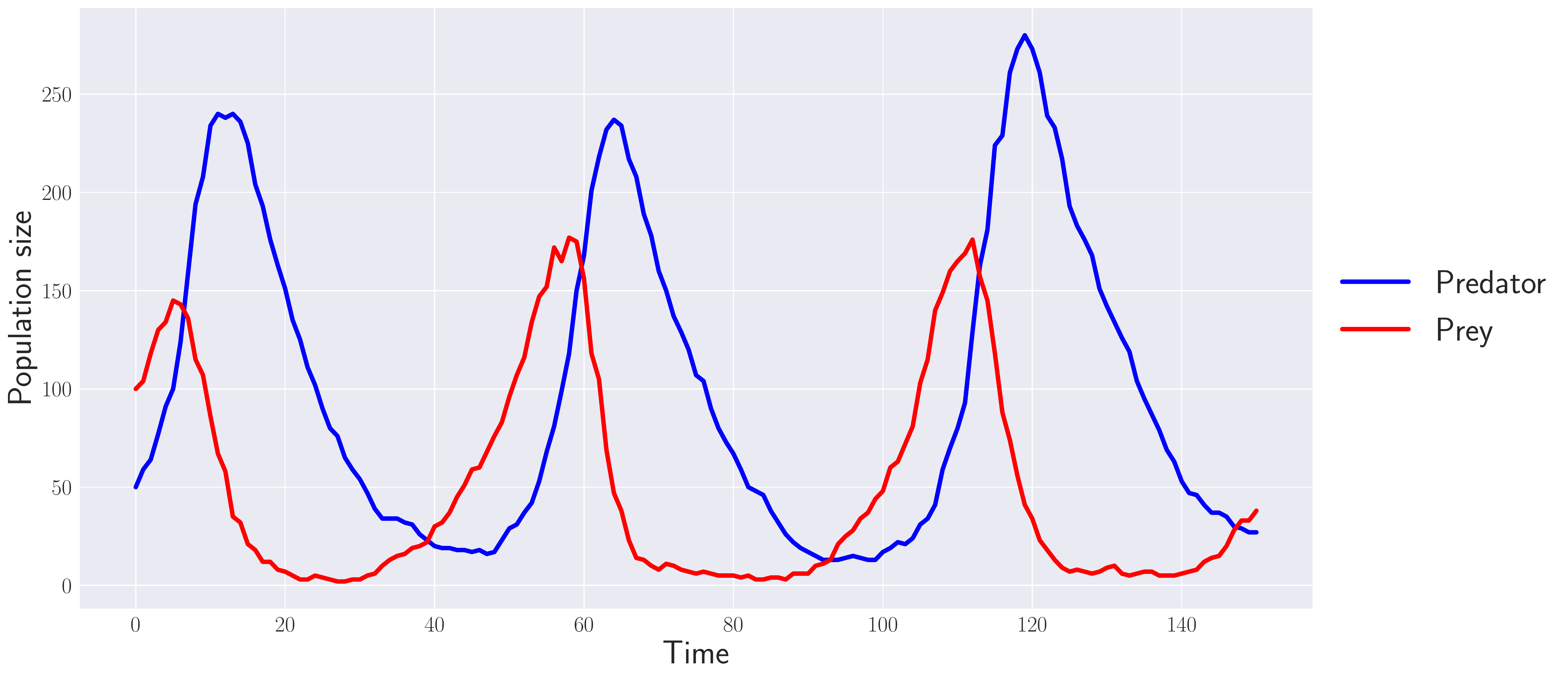}}
\caption{Lotka-Volterra: Simulated data set.}
\label{fig:LV_data}
\end{center}
\end{figure}

We considered the Markov-jump process version of the LV model presented in the supplementary material of \citet{papamakarios19snl}, including the model specification. The observed data, generated with ground-truth parameters $\theta_{\text{gt}} = [\log 0.01, \log 0.5, \log 1, \log 0.01]^T$,  is in Figure \ref{fig:LV_data}. For all parameters their prior is uniform  on $[-5, 2]$. 

\subsection{Summary statistics}

We used the same nine summary statistics as in \citet{papamakarios19snl}, that is: 

\begin{itemize}
    \item Mean and log variance of each time-series. 
    \item Auto-correlations of each time-series at time lags 1 and 2.
    \item Cross-correlation between the two time-series.
\end{itemize}
Before running the inference, a pilot-run procedure was used to make it possible to standardize the summary statistics so that, after standardization, the several components of each summary had a similar relevance. That is, in such a pilot run, we generated $1,000$ samples from the prior-predictive distribution: from these summaries, we then computed their trimmed means and trimmed standard deviations (we trimmed  the upper and lower 1.25\% of the distribution) to eliminate the effect of very extreme outliers. Afterwards, we ran each inference procedure using standardized summaries (both observed and simulated).

\subsection{Dealing with bad simulations}

Following \cite{papamakarios19snl} we used the Gillespie algorithm to simulate trajectories from the Lotka-Volterra model. The maximum allowed number of steps to advance the Gillespie simulation, for each given trajectory, was set to $10{,}000$. After $10{,}000$ steps we considered as the output of the simulator at $\theta$ the partially simulated trajectory and set to zero the remaining part of the path towards the simulation end-time. Thus, we did not remove parameter proposals that rendered poor simulations. 

\subsection{Posterior inference}

\begin{figure}[bt]
\begin{center}
\centerline{\includegraphics[width=1\columnwidth]{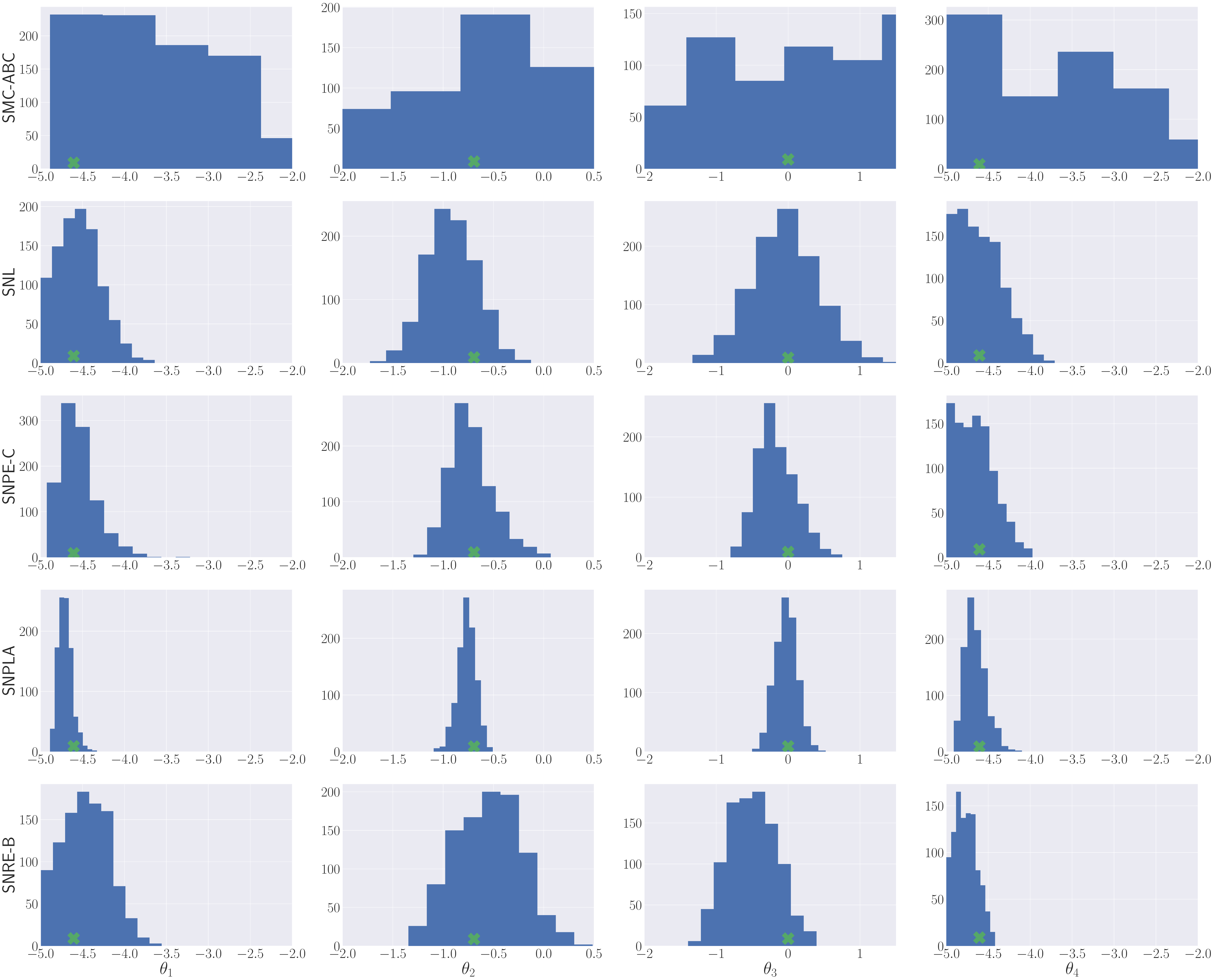}}
\caption{Lotka-Volterra: Samples from the resulting posterior approximations. The green marker shows the true parameter value. Results for one data set. }
\label{fig:LV_post_samples}
\end{center}
\end{figure}

Samples from the resulting posterior approximations for one of the runs are presented in Figure \ref{fig:LV_post_samples}. For the analysis, see the main paper.  

\subsection{Posterior predictive simulation}

\begin{figure}[tb]
\begin{center}
\centerline{\includegraphics[width=1\columnwidth]{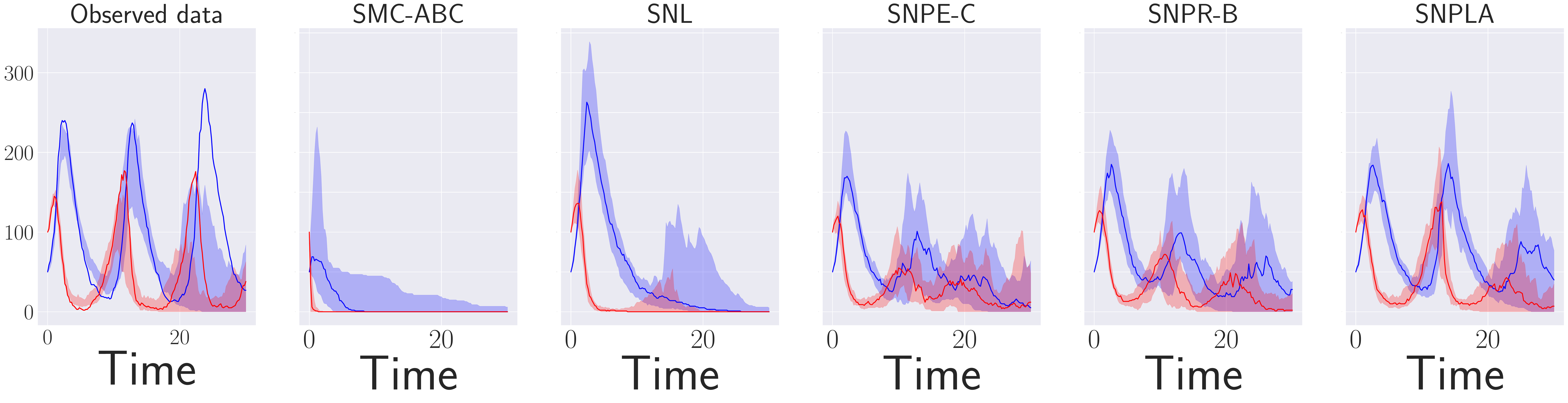}}
\caption{LV: Posterior predictive simulations. The solid lines show the median value over the attempts and the shaded areas show the range for the 25th and 75th percentile.}
\label{fig:LV_post_pred}
\end{center}
\end{figure}

The posterior predictive simulation are presented in Figure \ref{fig:LV_post_pred}. For the analysis, see the main paper. 

\section{Neural model: Hodgkin-Huxley model (HH)} \label{sec:HH}

\subsection{Model specification}

The HH equations \citep{hodgkin1952quantitative} model the dynamics of a neuron's membrane potential as a function of some stimulus (injected current) and a set of parameters. In our experiment, we use the same model formulation as in \cite{lueckmann2017flexible,papamakarios19snl}. A full description of the model can be found in the supplementary material for \cite{papamakarios19snl}. Our experimental setting is also similar to \cite{lueckmann2017flexible,papamakarios19snl}.

We considered 10 unknown parameters $\theta = [\log(\bar{g}_{Na})$, $\log(\bar{g}_K)$, $\log(g_{leak})$, $\log(E_{Na})$, $\log(-E_K)$, $\log(-E_{leak})$, $\log(\bar{g}_M)$, $\log(\tau_{max})$, $\log(V_t)$, $\log(\sigma)]$. The ground-truth values are: 
\begin{equation*}
  \begin{cases}
  \begin{split}
    &\bar{g}_{Na} = 200 \,\, (s/cm^2), \\
    &g_{leak} = 0.1 \,\, (s/cm^2), \\
    &E_K = -100 \,\, (mV), \\
    &\bar{g}_M = 0.07 \,\, (s/cm^2), \\
    &V_t = 60 \,\, (mV), \\ 
  \end{split}
\qquad
  \begin{split}
    &\bar{g}_K = 50 \,\, (s/cm^2), \\
    &E_{Na} = 50 \,\, (mV), \\
    &E_{leak} = -70 \,\, (mV), \\
     &\tau_{max} = 1000 \,\, (mV), \\
     &\sigma  = 1 \,\, (uA/cm^2). \\
  \end{split}
\end{cases}
\end{equation*}
We followed \cite{papamakarios19snl} and let the uniform prior for each unknown parameter  $\theta_i$ be 
\begin{align}
    \theta_i \sim U(\theta^{\star}_i - \log(2), \theta^{\star}_i + \log(1.5)),
\end{align}
where $\theta^{\star}_i$ is the ground-truth vale for $\theta_i$. 
We considered  $C$, $\kappa_{\beta n 1}$, $\kappa_{\beta n 2}$ as known, and these  were fixed at 
\begin{equation*}
    C = 1 \,\, (uF/cm^2), \qquad \kappa_{\beta n 1} =  0.5 \,\, (ms^{-1}), \qquad \kappa_{\beta n 2} =  40 \,\, (mV). 
\end{equation*}
We generated ``observed data'' from the HH model for $200 \,\, ms$ with a time-step of $0.025 \,\, ms$. We used the \texttt{Neuron} software \citep{carnevale2006neuron} to produce all model simulations.

\subsection{Summary statistics}

The likelihood $p(x | \theta)$ was defined on a set of 19 summary statistics that we computed from the voltage time-series produced by the \texttt{Neuron} simulator. We used the same 18 summary statistics as in \cite{papamakarios19snl}, and as an additional summary statistic, we included the number of spikes in the voltage time-series. The number of spikes in the data set is computed with the same methods as in \cite{lueckmann2017flexible}. Before running the inference, a pilot-run procedure was used to standardize the summary statistics using a whitening transform. Thus we used a similar standardization scheme for the summary statistics as in \cite{papamakarios19snl}.

\subsection{Posterior inference}

\begin{figure}[bt]
\begin{center}
\centerline{\includegraphics[width=1\columnwidth]{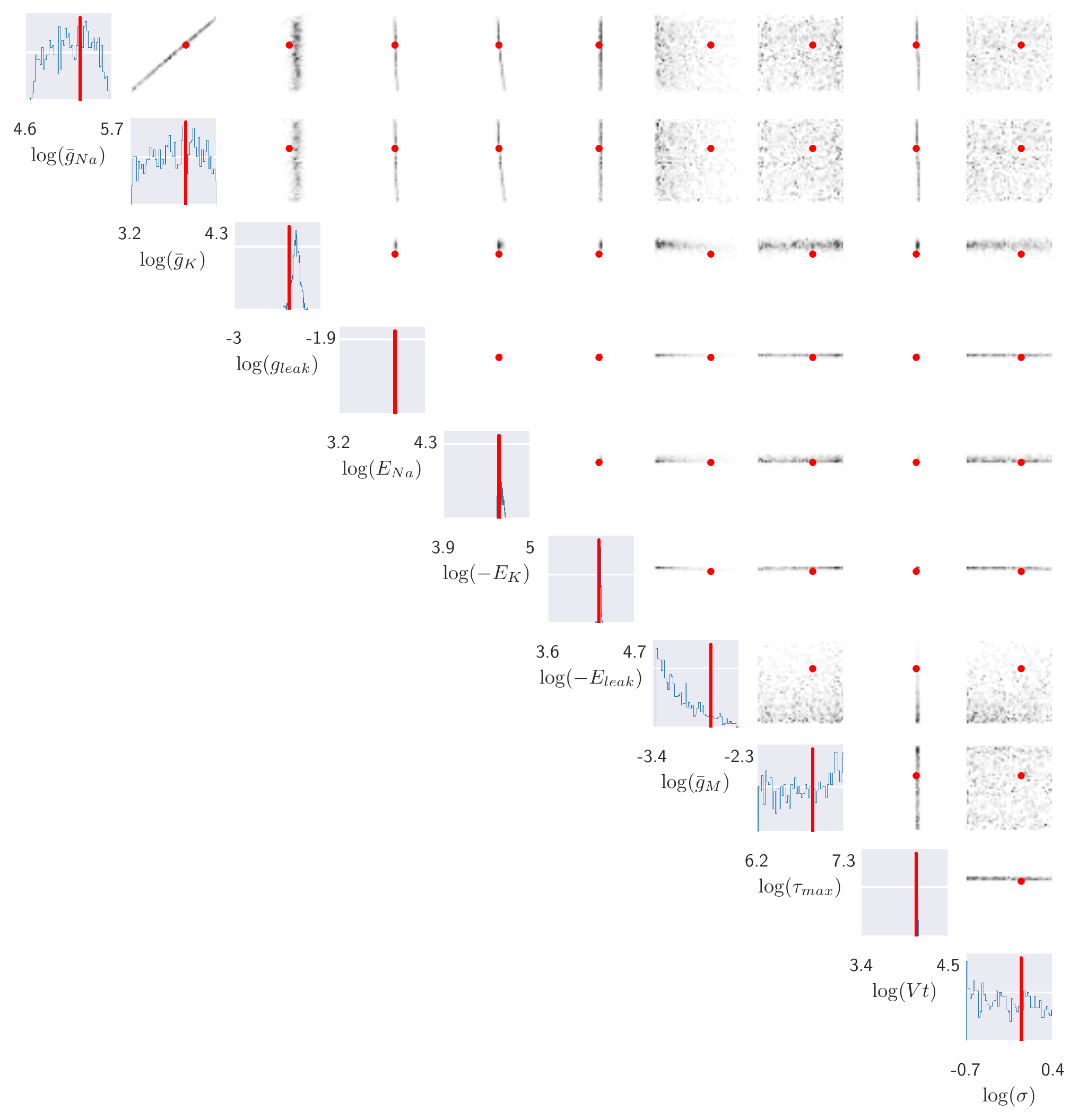}}
\caption{Hodgkin-Huxley: Samples from the posterior approximation for SNPE-C. Results for one data set.}
\label{fig:HH_snpe_c_post_samples}
\end{center}
\end{figure}

\begin{figure}[bt]
\begin{center}
\centerline{\includegraphics[width=1\columnwidth]{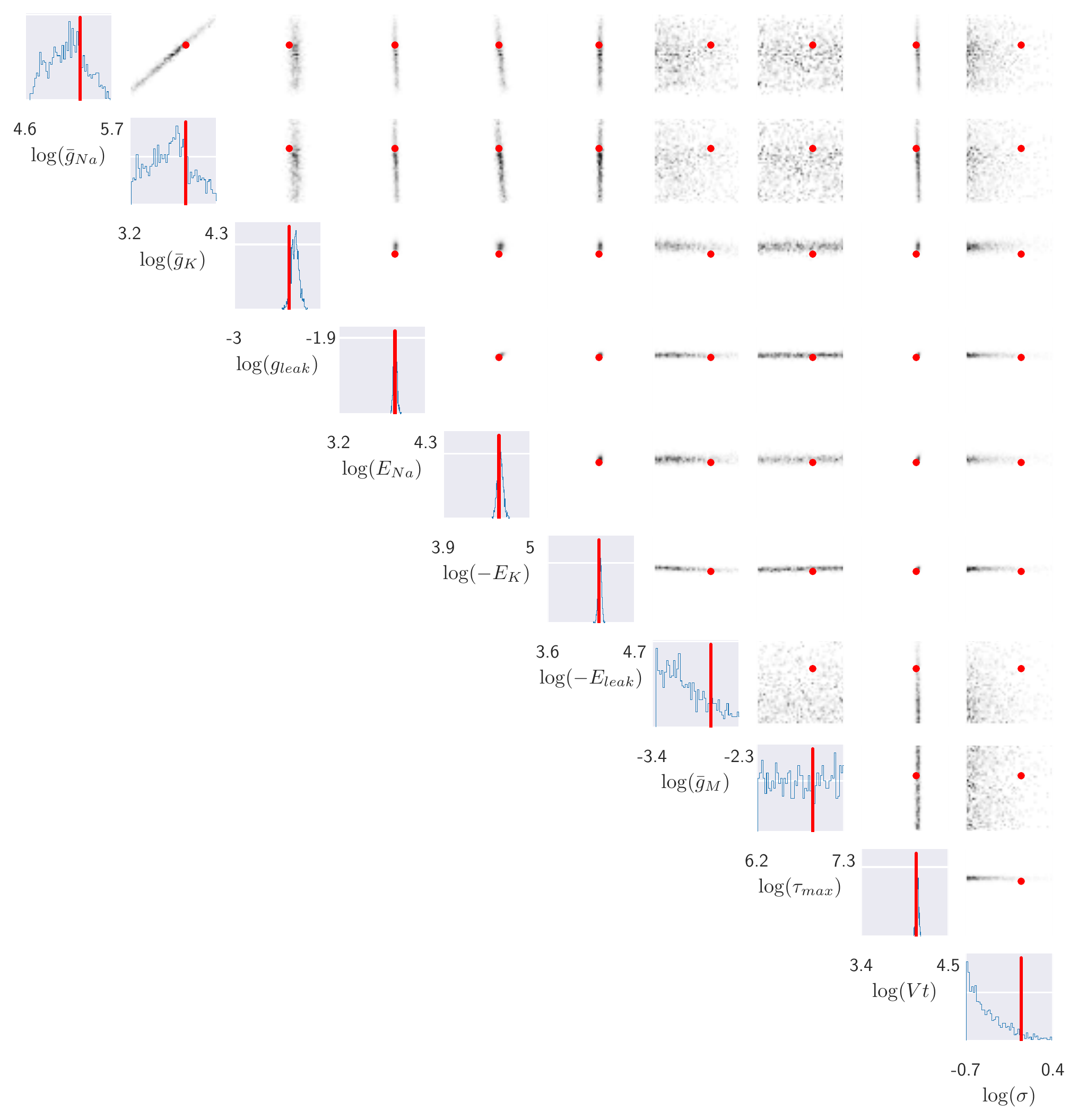}}
\caption{Hodgkin-Huxley: Samples from the posterior approximation for SNL. Results for one data set.}
\label{fig:HH_snl_post_samples}
\end{center}
\end{figure}

\begin{figure}[bt]
\begin{center}
\centerline{\includegraphics[width=1\columnwidth]{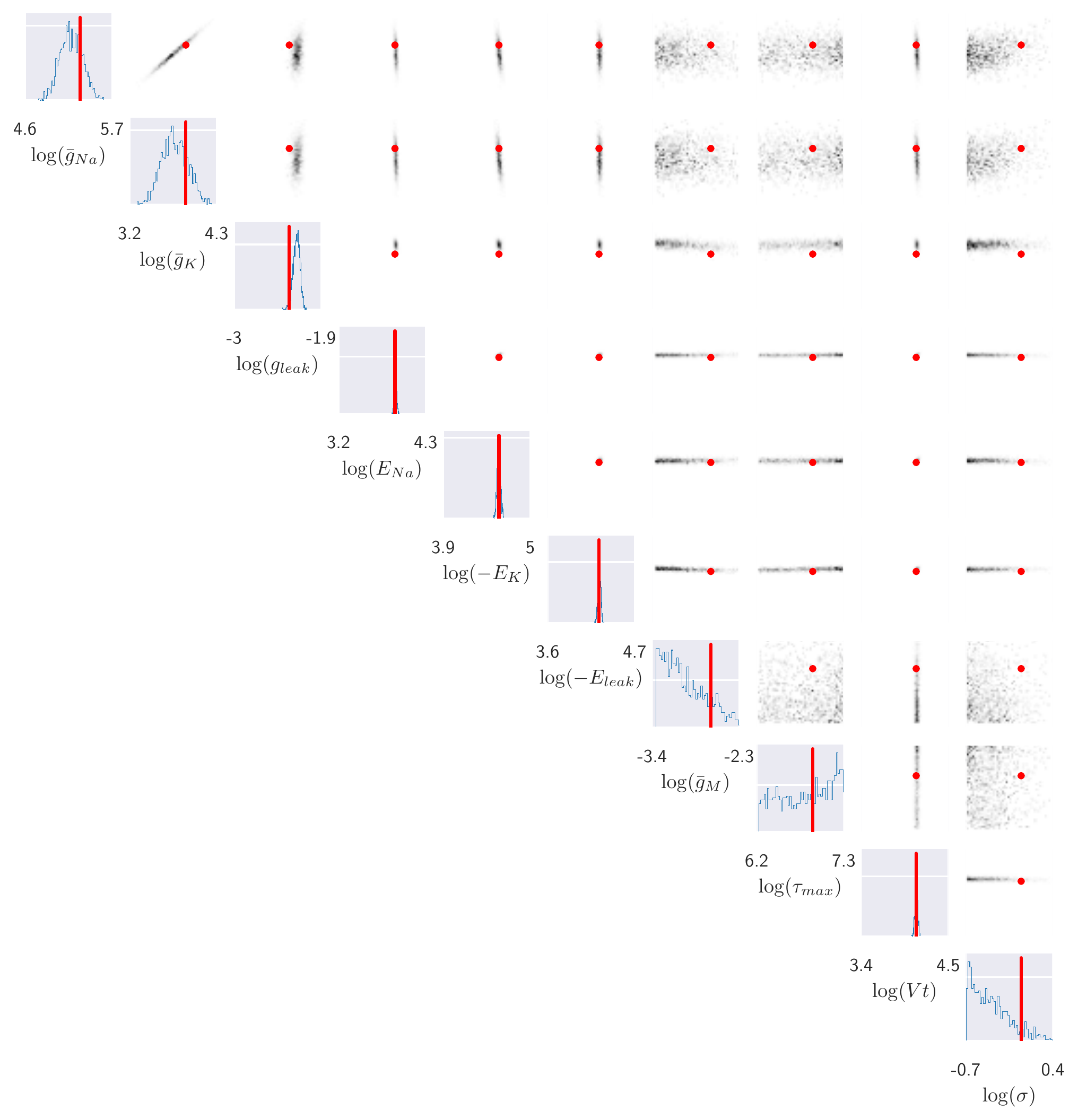}}
\caption{Hodgkin-Huxley: Samples from the posterior approximation for SNPLA. Results for one data set.}
\label{fig:HH_snpla_post_samples}
\end{center}
\end{figure}

The posterior samples from SNPE-C, SNL, and SNPLA are in Figures \ref{fig:HH_snpe_c_post_samples}, \ref{fig:HH_snl_post_samples}, and \ref{fig:HH_snpla_post_samples} respectively. For the analysis, see the main paper.  

\subsection{Posterior predictive simulations}

\begin{figure}[tb]
\begin{center}
\centerline{\includegraphics[width=1.\columnwidth]{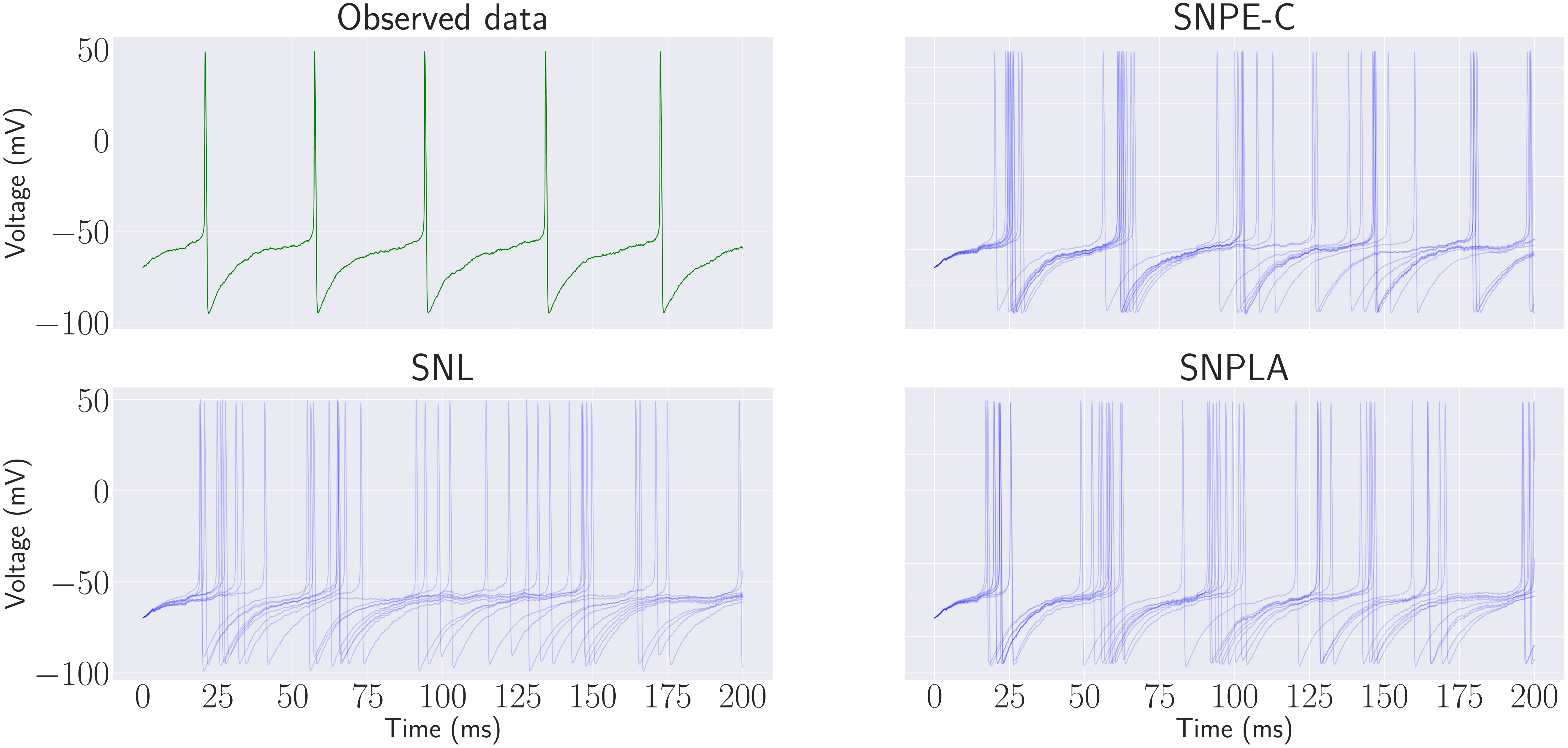}}
\caption{HH: Posterior predictive paths. Results for one data set.}
\label{fig:HH_post_pred}
\end{center}
\end{figure}

The posterior predictive simulations are presented in Figure \ref{fig:HH_post_pred}. For the analysis, see the main paper.  

\subsection{Prediction of the number of spikes}

\begin{figure}[ht]
\begin{center}
\centerline{\includegraphics[width=0.8\columnwidth]{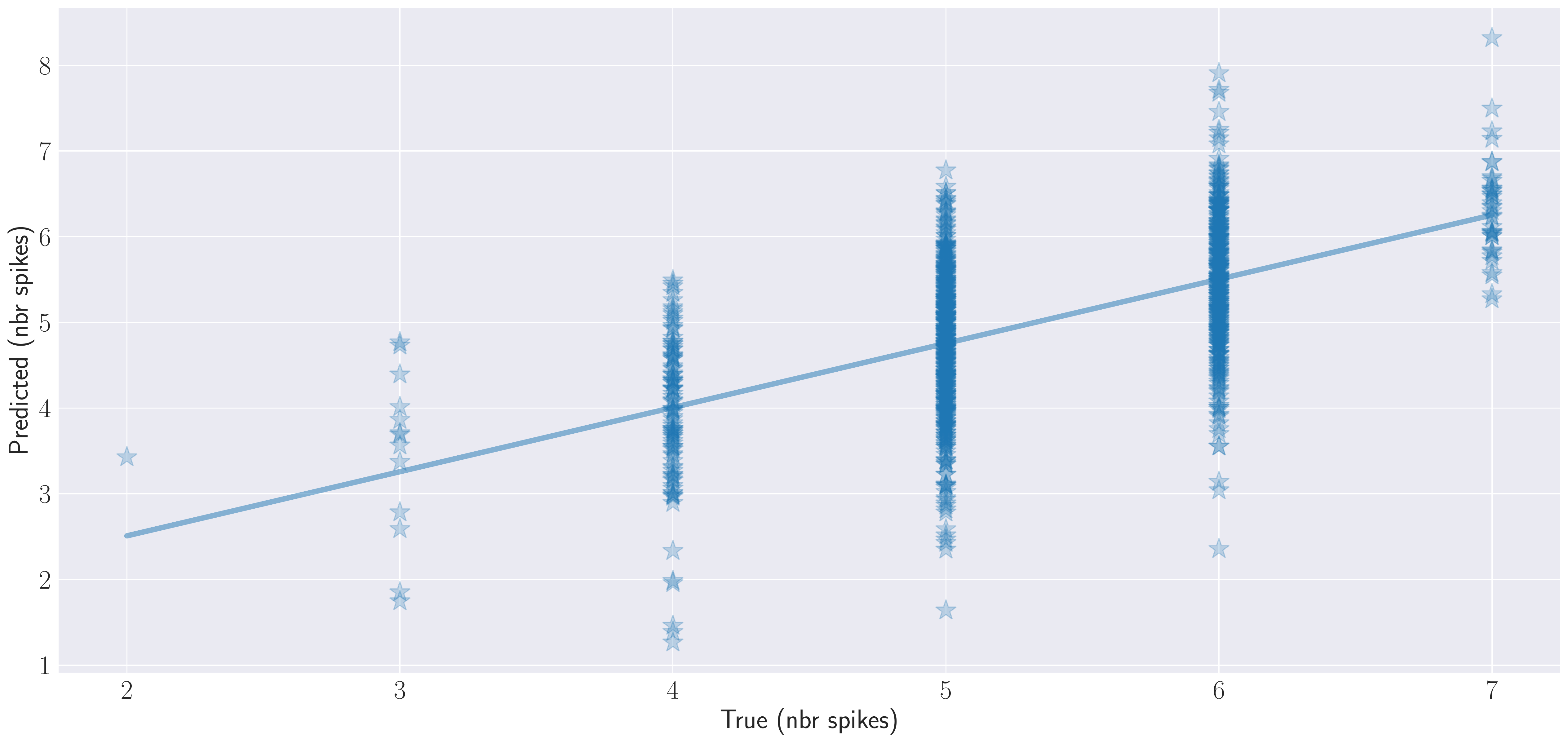}}
\caption{HH: Number of spikes prediction when using parameters from the resulting posterior. Results are based on a single data set. The solid line represents the line of best fit.}
\label{fig:HH_pred_num_spikes}
\end{center}
\end{figure}

\begin{figure}[ht]
\begin{center}
\centerline{\includegraphics[width=0.8\columnwidth]{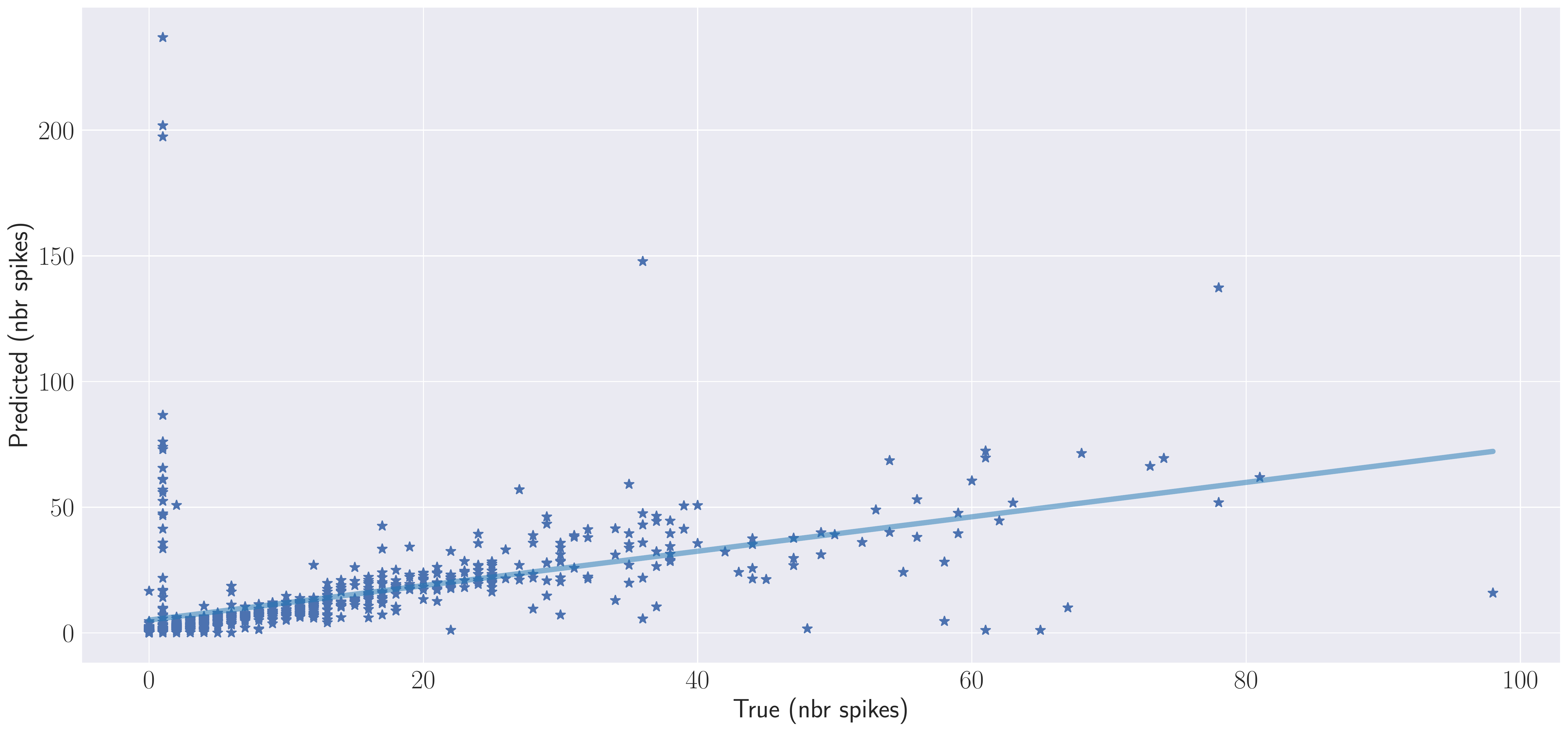}}
\caption{HH: Number of spikes prediction when using parameters generated from the prior. Results are based on a single data set. The solid line represents the line of best fit.}
\label{fig:HH_pred_num_spikes_prior}
\end{center}
\end{figure}

Figure \ref{fig:HH_pred_num_spikes} presents a comparison between the true number of spikes vs. the predicted number of spikes, for the case where parameters $\theta$ are generated from the resulting posterior distribution. We see that the predicted number of spikes in Figure \ref{fig:HH_pred_num_spikes} quite well corresponds to the true number of spikes. Figure \ref{fig:HH_pred_num_spikes_prior} shows the same analysis but for the case where where parameters $\theta$ are generated from the prior. However, the trained likelihood model occasionally predicted a very large number of spikes. Thus, in Figure \ref{fig:HH_pred_num_spikes_prior} we have removed 267 (out of 1000) cases where the  trained likelihood model predicted the number of spikes to be $> 1000$.   Unsurprisingly, compared to Figure \ref{fig:HH_pred_num_spikes} here we have that the predicted number of spikes matches the true number of spikes with higher uncertainty. 

\section{Training run-times}
The training run-times are presented in Table \ref{tab:runtime_training}.

\begin{table}[ht]
\caption{Median training run-time (in sec.).  MV Gaussian cases: (i) is ``five observations'', (ii) is ``summary statistics'', and (iii) is ``learnable summary statistics''. } 
\label{tab:runtime_training}
\centering
\begin{tabular}{lcccc} \toprule
 Experiment & SNL & SNPE-C & SNPLA & SNRE-B  \\ \midrule
MV-G (i)  & 7374 & 1179  & 1311 & 2676 \\   
MV-G (ii) & 13202 & 1696  & 1441 & 3340  \\ 
MV-G (iii)  &  7153   &  1803   & 1371 & 2586 \\
LV  &  6741   &   3628   &  3539   &  6167 \\
TM    &  3131    & 663 & 556 & 895  \\  
HH  &  42554   &   5240   &  4196   &  NA \\ \bottomrule
\end{tabular}
\end{table}

\section{Train/validation/test splits}

The train/validation/test splits used for the different experiments are presented in Table \ref{tab:train_test_val}. Regarding validation data: for SNL, SNPE-C, and SNRE-B, the validation data is set to a fraction $\mathrm{valfrac}$ of the training data. Thus, if we use $N$ samples to train the model, $N \times val-frac$ of these samples will be used for validation. This approach of splitting the training and validation data is also used when training the likelihood model for SNPLA. However, a somewhat different approach is used when training SNPLA's posterior model. Due to the \textit{simulation-on-the-fly} approach, for SNPLA's posterior model we  instead use $N_p \times \mathrm{valfrac}$ \textit{additional} simulations for validation purposes. 

\begin{table}[bt]
\caption{Train/validation/test splits for all experiments. $R$ is the number of rounds, $N$ is the number of model simulations per round, $N_p$ the number of total samples from the posterior model used to train SNPLA's posterior model, $\mathrm{valfrac}$ is the fraction of the training data used for validation,  $N_{test,post}$ is the number of posterior samples from the posterior approximation at each round, $N_{test,like}$ is the number of sample from the resulting likelihood model (only applicable to SNL and SNPLA). MV Gaussian cases: (i) is ``five observations'', (ii) is ``summary statistics'', and (iii) is ``learnable summary statistics''. }
\label{tab:train_test_val}
\centering
\begin{tabular}{lllllll} \toprule
Experiment & $R$ & $N$ & $N_p$ & $\mathrm{valfrac}$ & $N_{test,post}$ & $N_{test, like}$ \\ \midrule 
MV-G (i)  &  $10$ & $2{,}500$  & $40{,}000$ & $0.1$  & $1{,}000$  &  NA    \\
MV-G (ii)  &  $10$ & $2{,}500$  & $10{,}000$ & $0.1$  & $1{,}000$  &  $1{,}000$      \\
MV-G (iii)  &  10 & $2{,}500$  & $40{,}000$ & 0.1  & $1{,}000$  &  NA      \\
TM  &  10 & $1{,}000$  & $60{,}000$ & $0.1$  & $1{,}000$  &  $1{,}000$      \\ 
LV  &  5 & $1{,}000$  & $10{,}000$ & $0.1$  & $1{,}000$  &  NA    \\ 
HH  & 12 & $2{,}000$  & $10{,}000$ & $0.1$  & $1{,}000$  &  NA    \\  \bottomrule
\end{tabular}
\end{table}

\section{Hyper-parameter settings} \label{sec:hyper_params}

The hyper-parameter settings, for all experiments, are in Tables \ref{tab:hp_snl}-\ref{tab:hp_snpla}. 
Regarding the learn-rate settings: $0.0005$ is the default learn-rate used in \texttt{sbi}, and $0.001$ is the default in the Adam optimizer found in \texttt{PyTorch}. 
We decreased the learn rate for SNPLA's posterior model using the \texttt{PyTorch} function \\ \texttt{torch.optim.lr\_scheduler.ExponentialLR} with multiplicative factor $\gamma_p$, see Table \ref{tab:hp_snpla} (for details on how the multiplicative factor $\gamma_p$ is used, see the documentation for \texttt{torch.optim.lr\_scheduler.ExponentialLR}). 

\renewcommand*\footnoterule{}

\begin{table}[!ht]
\begin{minipage}{\textwidth}
        \centering
        \caption{SNL: Hyper-parameter-setting for the different experiments for SNL. $lr$ is the learn rate.  MV Gaussian cases: (i) is ``five observations'', (ii) is ``summary statistics'', and (iii) is ``learnable summary statistics''. }
        \label{tab:hp_snl}
        \begin{tabular}{ll} \toprule
                Experiment    & $lr$ \\ \midrule
        MV-G (i)   &  $0.0005$    \\
        MV-G (ii)   &     $0.0005$   \\
        MV-G (ii)   &     $0.0005$    \\
        TM          &   $0.0005$    \\ 
        LV          &  $0.0005$\footnote{From a numerical-stability standpoint we found it for this case to be beneficial to exponentially decrease the learn rate with a decay rate of $0.98$}       \\ 
        Hodgkin-Huxley          &   $0.0005$     \\ \bottomrule 
        \end{tabular}
\end{minipage}

\end{table}

\begin{table}[!ht]
        \centering
        \caption{SNPE-C: Hyper-parameter-setting for the different experiments for SNPE-C.  $lr$ is the learn rate. MV Gaussian cases: (i) is ``five observations'', (ii) is ``summary statistics'', and (iii) is ``learnable summary statistics''.}
        \label{tab:hp_snpe_c}
        \begin{tabular}{ll} \toprule
          Experiment          & $lr$ \\ \midrule
        MV-G (i)  &     $0.0005$     \\
        MV-G (ii)  &      $0.0005$    \\
        MV-G (iii) &     $0.0005$     \\
        TM          &      $0.0005$    \\ 
        LV          &    $0.0005$      \\ 
        HH          &    $0.0005$      \\ \bottomrule 
        \end{tabular}
\end{table}

\begin{table}[!ht]
        \centering
        \caption{SNRE-B: Hyper-parameter-setting for the different experiments for SNPE-B.  $lr$ is the learn rate. MV Gaussian cases: (i) is ``five observations'', (ii) is ``summary statistics'', and (iii) is ``learnable summary statistics''.}
        \label{tab:hp_snre_B}
        \begin{tabular}{ll} \toprule
          Experiment          &  $lr$ \\ \midrule
        MV-G (i)  &   $0.0005$     \\
        MV-G (ii)  &    $0.0005$      \\
        MV-G (iii) &    $0.0005$      \\
        TM          &    $0.0005$      \\ 
        LV          &    $0.0005$      \\ 
        HH          &     NA     \\ \bottomrule 
        \end{tabular}
\end{table}

\begin{table}[!ht]
        \centering
        \caption{SNPLA: Hyper-parameter-setting for the different experiments for SNPLA.  $lr_L$ is the learn rate for the likelihood model, $lr_P$ is the learn rate for the posterior model, $\gamma_P$ is the multiplicative factor of the decrease for the learn rate of the posterior model, and $\lambda$ is the exponential decrease rate for the prior. MV Gaussian cases: (i) is ``five observations'', (ii) is ``summary statistics'', and (iii) is ``learnable summary statistics''.}
        \label{tab:hp_snpla}
        \begin{tabular}{lllll} \toprule
              Experiment       & $lr_L$ & $lr_P$ & $\gamma_P$ & $\lambda$  \\ \midrule
        MV-G (i)  & $0.001$  & $0.002$  &  $0.95$    &  $0.7$ \\
        MV-G (ii)  & $0.001$  & $0.002$  &  $0.95$    &  $0.7$  \\
        MV-G (iii)  & $0.001$  & $0.002$  &  $0.95$    &  $0.7$    \\
        TM          &  $0.001$  & $0.001$  &  $0.9$    &  $0.7$  \\ 
        LV          & $0.001$  & $0.001$  &  $0.9$    &  $0.9$     \\ 
        HH          & $0.001$  & $0.001$  &  $0.95$    &  $0.8$    \\ \bottomrule 
        \end{tabular}
\end{table}

\section{Sensitivity analysis: hyper-parameter ranges}

The Hyper-parameter-ranges used for the sensitivity analysis are presented in Table \ref{tab:hp_ranges}. 

\begin{table}[!ht]
        \centering
        \caption{Hyper-parameter-ranges: SNL shows the Hyper-parameter-ranges used for MV-G ((i), (ii), (iii)) and TM, SNL-LV shows the Hyper-parameter-ranges used for LV. SNPE-C, SNPLA, and SNRE-B shows the Hyper-parameter-ranges used for MV-G (i,ii,iii), TM, and LV  }
        \label{tab:hp_ranges}
        \begin{tabular}{lllllll} \toprule
        Method  & $lr$ & $\gamma_{lr}$ & $lr_L$ & $lr_P$ & $\gamma_P$ & $\lambda$  \\ \midrule
        SNL  & $[10^{-4}, 10^{-2}]$ & NA & NA & NA  &  NA    &  NA \\
        SNL-LV & $[10^{-4}, 10^{-2}]$ & $[0.9,0.999]$ & NA & NA  &  NA    &  NA  \\
        SNPE-C & $[10^{-4}, 10^{-2}]$ & NA & NA & NA  &  NA    &  NA  \\
        SNPLA & NA & NA & $[10^{-4}, 10^{-2}]$ & $[10^{-4}, 10^{-2}]$  &  $[0.8,0.999]$    &  $[0.65, 0.95]$ \\
        SNRE-B & $[10^{-4}, 10^{-2}]$ & NA & NA & NA  &  NA    &  NA  \\ \bottomrule 
        \end{tabular}
\end{table}

\section{SNPLA for different values of $\lambda$}

\begin{figure}[ht]
\begin{center}
\centerline{\includegraphics[width=0.8\columnwidth]{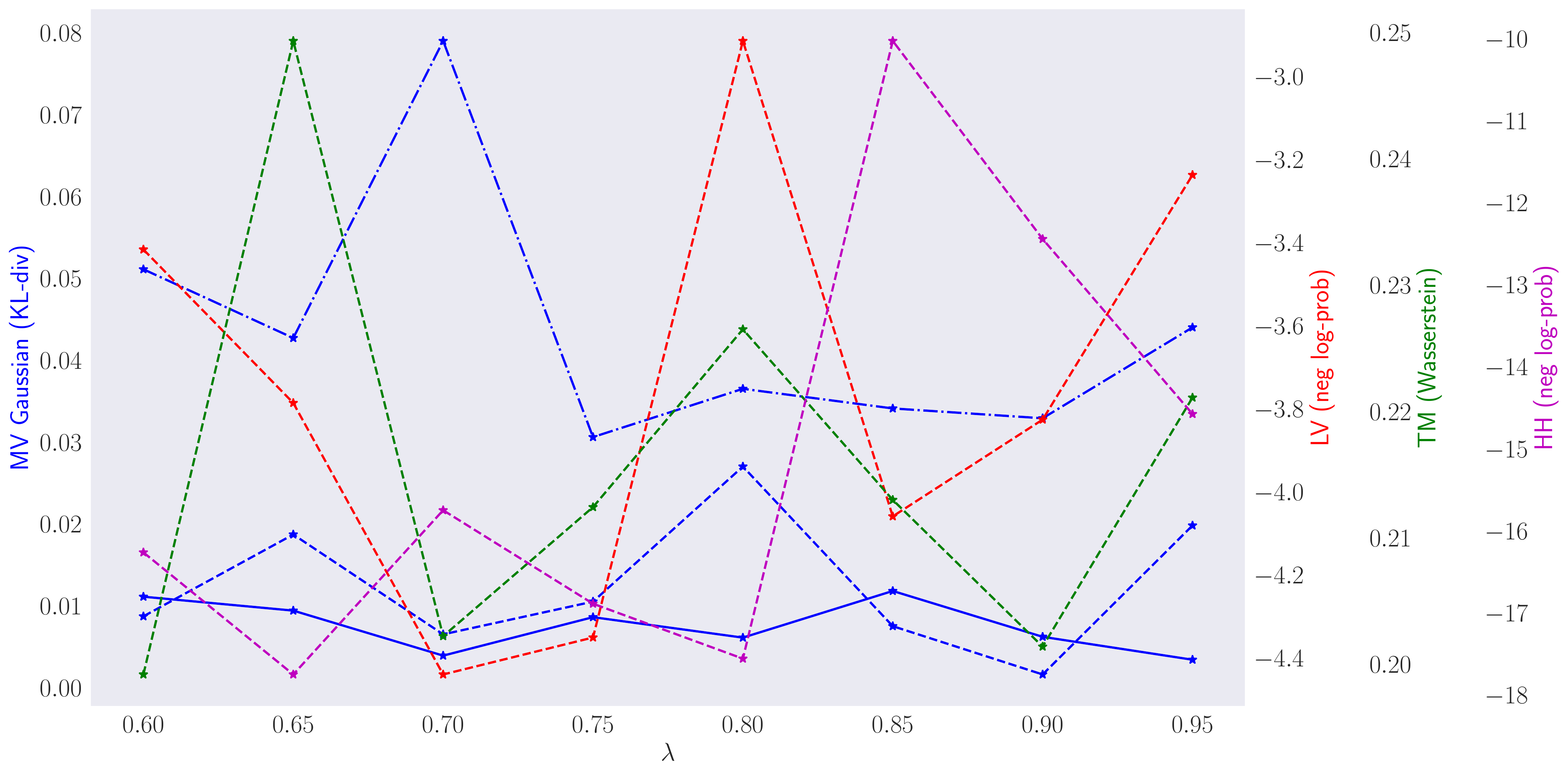}}
\caption{Results for different $\lambda$ values.}
\label{fig:diff_lambda}
\end{center}
\end{figure}

Figure \ref{fig:diff_lambda} presents the performance measures that we obtain when running SNPLA with different $\lambda$ values. We conclude that the difference in performance when changing $\lambda$ is moderate, and that for all cases we obtain a resulting performance measure indicating adequate posterior approximation.    

\end{document}